\documentclass[submission,copyright]{eptcs}
\providecommand{\event}{5th Annual International Conference on Applied Category Theory (ACT 2022)} 
\usepackage{breakurl}             
\usepackage{underscore}           
\usepackage[frozencache,cachedir=_minted-output]{minted}

\usepackage{algorithm}
\usepackage{algorithmicx}
\usepackage{algpseudocode}
\usepackage{amsmath,amsfonts,amssymb,amsthm,bm}
\usepackage{mathtools}
\usepackage{cancel}

\newtheorem{condition}{Condition}
\newtheorem{definition}{Definition}
\newtheorem{lemma}{Lemma}
\newtheorem{proposition}{Proposition}
\newtheorem{theorem}{Theorem}
\newtheorem{conjecture}{Conjecture}

\usepackage{booktabs}
\usepackage{stmaryrd}           

\usepackage{tikzit}             

\tikzstyle{Medium Box}=[fill=white, draw=black, shape=rectangle, tikzit category=Box, minimum width=0.75cm, minimum height=0.5cm]
\tikzstyle{dot}=[fill=black, draw=black, shape=circle, tikzit shape=circle, inner sep=0.4mm]
\tikzstyle{triangle}=[fill=white, draw=black, shape=regular polygon, regular polygon sides=3, text width=2mm, inner sep=0.5mm, outer sep=0mm, shape border rotate=180]
\tikzstyle{medium box 90}=[fill=white, draw=black, shape=rectangle, tikzit category=Box, minimum width=0.75cm, minimum height=1cm]
\tikzstyle{small box}=[fill=white, draw=black, shape=rectangle, tikzit category=Box, minimum height=0.4cm, minimum width=0.3cm]
\tikzstyle{large rounded box}=[fill=none, draw=black, shape=rectangle, tikzit category=Box, minimum height=2.5cm, minimum width=3.5cm, rounded corners=0.5cm]
\tikzstyle{medium rounded box}=[fill=none, draw=black, shape=rectangle, tikzit category=Box, minimum width=1cm, minimum height=0.75cm, rounded corners=0.05cm]
\tikzstyle{small rounded box}=[fill=none, draw=black, shape=rectangle, tikzit category=Box, minimum width=0.5cm, minimum height=0.4cm, rounded corners=0.05cm]
\tikzstyle{medium dashed box}=[fill=none, draw=black, shape=rectangle, dashed, tikzit category=Box, minimum width=1.42cm, minimum height=0.8cm, rounded corners=0.05cm]

\tikzstyle{new edge style 0}=[->]

\usepackage{pifont}
\usepackage{graphicx}
\usepackage{subcaption}
\usepackage{hyperref}

\newcommand{\cmark}{\ding{51}}
\newcommand{\xmark}{\ding{55}}

\newcommand{\prob}[1]{\mathbb{P}\: #1}
\newcommand{\p}[1]{\mathbf{P}\: #1}
\newcommand{\freeCat}[1]{\mathbf{Free}(#1)}
\newcommand{\kl}[2]{\text{KL}(#1 || #2)}
\newcommand{\expect}[2]{\mathbb{E}_{#1} \left[ #2 \right]}
\newcommand{\expectint}[3]{\int\!{d #2}\: #1 \: #3}

\title{A Probabilistic Generative Model of Free Categories}

\author{
Eli Sennesh
\institute{Khoury College of Computer Sciences}
\institute{Northeastern University\\
Boston, Massachusetts, USA}
\email{sennesh.e@northeastern.edu}
\and
Tom Xu
\institute{School of Computing\\
The Australian National University\\
Canberra, Australia}
\email{tom.xu@anu.edu.au}
\and
Yoshihiro Maruyama
\institute{School of Computing\\
The Australian National University\\
Canberra, Australia}
\email{yoshihiro.maruyama@anu.edu.au}
}

\def\titlerunning{A Probabilistic Generative Model of Free Categories}
\def\authorrunning{E. Sennesh, T. Xu, \& Y. Maruyama}
\begin{document}
\maketitle

\begin{abstract}
Applied category theory has recently developed libraries for computing with morphisms in interesting categories, while machine learning has developed ways of learning programs in interesting languages. Taking the analogy between categories and languages seriously, this paper defines a probabilistic generative model of morphisms in free monoidal categories over domain-specific generating objects and morphisms. The paper shows how acyclic directed wiring diagrams can model specifications for morphisms, which the model can use to generate morphisms. Amortized variational inference in the generative model then enables learning of parameters (by maximum likelihood) and inference of latent variables (by Bayesian inversion). A concrete experiment shows that the free category prior achieves competitive reconstruction performance on the Omniglot dataset.
\end{abstract}

\section{Introduction}
\label{sec:intro}
\vspace{-0.5em}

Applied category theory has recently developed software libraries for representing and computing with morphisms in various categories, with a particular focus on (symmetric) monoidal categories and diagrammatic reasoning (e.g. \cite{DeFelice2020,Halter2020,Patterson2021,Patterson2021b}). Categories have also emerged as a common representation for computer programs in different languages \cite{sorensen2006lectures,Elliott2017}.  In parallel, the field has worked to provide categorical semantics and interpretations for a variety of machine learning (ML) and artificial intelligence (AI) building blocks, including neural networks \cite{Fong2019,Fong2019a,Cruttwell2021}, and probabilistic inference \cite{giry1982categorical,Culbertson2014,Cho2019,Fritz2020}.

The intersection of learning and compositional reasoning has challenged the cognitive sciences (AI, ML, cognitive science, etc.) from the beginning. Theoretical arguments suggest that a generally intelligent agent \emph{ought} to reason compositionally \cite{Phillips2010,marcus2018algebraic}, to employ something like computable programs as models of the world \cite{SOLOMONOFF19641}, and to prefer simpler programs to more complex ones \cite{Vitanyi2000}. Evidence from a variety of experiments \cite{Lake2015,Ho2018,Grant2019,Lake2019,Zhou2021} suggests that abductive learning of program-like causal models provides a domain-general foundation for human intelligence.

Most process models of cognition spring from a limited selection of metaphors for how the mind works. For instance, cognitive scientists continue to debate whether the mind builds concepts by means of a ``language of thought'' \cite{Piantadosi2016a,Overlan2017,Romano2018} or convex geometric spaces \cite{gardenfors2004conceptual}. Some cognitive neuroscientists find tentative support for both ``map-like'' and ``sentence-like'' processes in the brain \cite{Frankland2020} while others argue for a more connectionist ``direct fit'' \cite{Hasson2020}. Machine learning scientists tackling the learning and synthesis of task-specific policies or programs tend to employ either Bayesian inference over a context-free grammar of programs, reinforcement learning of neural network policies, or both \cite{Parisotto2017,Valkov2018,Nye2020}.

Category theory readers will notice that program learning tasks, via their context-free grammars, work with the compositional structure of 
operads \cite{Hermida1998,Fong2019b}. The connection suggests that categories, too, could provide a setting for learning and inference. This paper thus suggests an additional alternative to the options of neural networks and grammatical programs: a probabilistic generative model (PGM) of morphisms in a free monoidal category. This \emph{free category prior} is parameterized by generating objects and morphisms (to construct the free category) and wiring diagrams (to sample specific morphisms). An experiment shows that the model's software implementation can learn to compose deep generative models to achieve competitive performance in reconstruction evaluation data.

\vspace{-1em}
\paragraph{Outline}
Section~\ref{sec:background} summarizes the categorical background for the rest of the paper: our chosen setting for categorical probability in Section~\ref{subsec:Markov}, and the operad of wiring diagrams for use as specifications in Section~\ref{subsec:wiringoperad}. Section~\ref{sec:categoricalpgm} then defines the necessary categorical machinery for generative modeling, presents the free category prior, and demonstrates its basic properties. Section~\ref{sec:software} describes its software implementation, describes amortized variational inference (Section~\ref{subsec:amortizedinf}), and presents a basic experiment (Section~\ref{subsec:experiments}) showing that fitting the free category model to data yields competitive performance in a generative modeling task. Section~\ref{sec:discussion} discusses extensions and further applications.

\vspace{-1em}
\section{Categorical background}
\label{sec:background}

\subsection{Markov categories}
\label{subsec:Markov}
In this paper we build upon Markov categories, a formalism presented by Fritz~\cite{Fritz2020} following a series of works by Golubtsov~\cite{golubtsov1999axiomatic} and Cho and Jacobs~\cite{Cho2019}. 
It allows us to represent fundamental concepts from probability theory (such as conditioning, causality, almost surety, etc.)  in purely categorical terms. 
In particular we work in \textbf{QBS} (i.e., the category of quasi-Borel spaces) by Heunen et al. \cite{Heunen2017}.
In the following we use the notation $\fatsemi$ and $\odot$ throughout for sequential and monoidal composition. 

\begin{definition}[Markov Category]
\label{def:markovcat}
A Markov category \(\mathcal{C}\) is a symmetric monoidal category in which every object \(X \in \mathcal{C}\) is equipped with a commutative comonoid structure given by a comultiplication \(\textbf{copy}_X: X \rightarrow X \odot X\) and a counit \(\textbf{del}_X: X \rightarrow I\).
\end{definition}

Modeling probability theory in a Markov category involves the following steps:
\begin{itemize}
    \item Identify a base Markov category \(\mathcal{C}\) (usually a category of certain type of measurable spaces).
    \item Identify a monad (usually Giry/probability monad) represented by an endofunctor \(\prob{}: \mathcal{C} \rightarrow \mathcal{C}\).
    \item Take the Kleisli category \(Kl(\prob{})\) of the monad \(\prob{}\).
\end{itemize}
Corollary 3.2 in \cite{Fritz2020} says that if \(\prob{}\) is a symmetric monoidal affine monad then the Kleisli category \(Kl(\prob{})\) is again a Markov category. In particular, the Giry monad is symmetric monoidal and affine.

Typically Markov categories consist of certain measurable spaces as objects and Markov kernels as morphisms. In \textbf{QBS}, by contrast, each object consists of a pair of a sample space $X$ and a collection of $X$-valued random variables $M_{X} \subseteq[\mathbb{R} \rightarrow X]$ (which must be closed under certain conditions) and the morphisms are random functions between sample spaces that extend the random variables in the domain to random variables in the codomain. Random variables are thus fundamental, rather than being derived from a $\sigma$-algebra of measurable subsets. They then derive their randomness from probability measures in $\mathbb{R}$, as we will discuss below.



\begin{definition}[Category of quasi-Borel spaces]
\label{def:qbs}
A quasi-Borel space is a set $X$, which we call the \emph{sample space}, together with a set of functions $M_X \subseteq [\mathbb{R} \rightarrow X]$ that
\begin{itemize}
    \item Contains all constant functions: \(\alpha \in M_{X}\) if \(\alpha: \mathbb{R}\rightarrow X\) is constant;
    \item Is closed under pre-composition: \(\alpha \circ f \in M_{X}\) if \(\alpha \in M_X\) and \(f: \mathbb{R}\rightarrow \mathbb{R}\) is measurable; and
    \item Is closed under countable mixtures: with respect to a partition of \(\mathbb{R} \) into a  disjoint union of countably many Borel sets i.e. \(\mathbb{R}=\biguplus_{i \in \mathbb{N}} S_{i}\), and \(\alpha_{1}, \alpha_{2}, \ldots \in M_{X}\) then the mixed random variable \(\beta\) is in \(M_{X}\) where \(\beta(r) = \alpha_i(r)\) for \(r\in S_i\).
\end{itemize}
Morphisms \((X, M_X) \rightarrow (Y, M_Y)\) between quasi-Borel spaces consist of  functions \(f: X\rightarrow Y\) such that \(f\circ \alpha \in M_Y\) for all \(\alpha \in M_X\).

Together these form the category of Quasi-Borel Spaces and measurable functions between them (\textbf{QBS}), with function composition as composition of morphisms and identity functions as identity morphisms.
\end{definition}





To study probability theory via quasi-Borel spaces, there is a notion of probability measure on them.
\begin{definition}[Probability measure on quasi-Borel space.]
\label{def:probmeas}
A probability measure on a quasi-Borel space \((X, M_X)\) is a pair \((\alpha, \mu)\) where \(\alpha \in M_X\) and \(\mu\) is a probability measure on \(\mathbb{R}\). Here \(\mu\) is a probability measure from the standard measure theory.
\end{definition}
The above definition reflects the usual notion of a probability measure on a measurable space which is induced via pushing-forward through a random variable (a measurable function) from the sample space with a probability measure. Another way of phrasing the notion of probability measure is that with a source of randomness from \(\mathbb{R}\) and a distribution \(\mu\) on \(\mathbb{R}\), we observe the effect of the process \(\alpha \in M_X\) in \(X\). From this notion of probability measure, there is generalized version of Giry monad in \textbf{QBS} which is referred as the probability monad.

\begin{lemma}[Probability monad in \textbf{QBS}, due to Heunen et al.~\cite{Heunen2017}]
\label{lem:probMonad}
The set of all probability measures on a quasi-Borel space naturally forms a quasi-Borel space (see Heunen et al.~\cite{Heunen2017}). The endofunctor \(\prob\) sending a quasi-Borel \(X\) to the space \(\prob{X}\) of probability measures on \(X\) (quotienting out a suitable notion of equivalent measures) gives rise to a commutative monad in \textbf{QBS} with Kleisli composition $\fatsemi_{\prob{}}: \mathbf{QBS}(A, \prob{B}) \times \mathbf{QBS}(B, \prob{C}) \rightarrow \mathbf{QBS}(A, \prob{C})$ and unit natural transformation $\eta: \mathbb{I} \rightarrow \prob{}$ from the identity functor $\mathbb{I}$.
\end{lemma}

\vspace{-1em}
\subsection{Wiring diagram operads and algebras over them}
\label{subsec:wiringoperad}
Rupel and Spivak~\cite{Rupel2013} introduced wiring diagrams as a formal graphical language to represent the structure of composite processes. Despite their geometric presentation, wiring diagrams should be understood as combinatorial entities. Every wiring diagram is a combinatorial scheme of composing blank boxes and directed wires, and Patterson et al.~\cite{Patterson2021} showed that wiring diagrams provide a combinatorial normal form for morphisms in SMCs. This makes wiring diagrams suitable to be represented in a data structure. This paper employs acyclic wiring diagrams as a syntax for specifying morphisms in categories.

Mathematically, wiring diagrams are expressed via the notion of a \textit{symmetric colored operad}. Loosely speaking, a colored operad is a category except that the hom-sets are allowed to go from a one finite set of objects to another finite set of objects. Operads capture algebraic structures in SMCs. We utilize the specific definition of an operad found in Spivak~\cite{Spivak2013}. In short, an operad \(\mathcal{O}\) is encoded by two pieces of data: structure and laws. The structure of \(\mathcal{O}\) consists of a) objects; b) morphisms; c) specified identity morphisms on objects; and d) a composition formula for morphisms. The structure is required to satisfy an identity law and an associativity law. We give a quick overview of the \(\tau\)-typed operad \(\mathcal{W_\tau}\) of acyclic wiring diagrams for the convenience of readers. The full definition can be found in Patterson~\cite{Patterson2021}.

\begin{definition}[Operad of acyclic wiring diagrams]
Let \(\tau\) be a set. A \(\tau\)-typed set is an object in the slice category \(\mathtt{Set}/\tau\). The operad \(\mathcal{W_\tau}\) has the following structure:
\begin{itemize}
    \item \textit{Objects}. An object (\textbf{box}) \(\mathbf{t}\) of \(\mathcal{W_\tau}\) is a pair of \(\tau\)-typed finite sets. Pictorially \(\mathbf{t}=(\mathbf{t}_{-}, \mathbf{t}_{+})\) is a blank box with a finite number of \(\tau\)-typed input and output wires \footnote{The wires are directed from left to right.}.
    \begin{center}
        \scalebox{1.4}{\begin{tikzpicture}
	\begin{pgfonlayer}{nodelayer}
		\node [style=medium rounded box] (0) at (0.25, -0.25) {};
		\node [style=none] (12) at (2.25, 0.25) {};
		\node [style=none] (13) at (2.25, 0) {};
		\node [style=none] (14) at (2.25, -0.75) {};
		\node [style=none] (16) at (1.25, 0.25) {};
		\node [style=none] (17) at (1.25, 0) {};
		\node [style=none] (18) at (1.25, -0.75) {};
		\node [style=none] (19) at (-0.75, 0.25) {};
		\node [style=none] (20) at (-0.75, 0) {};
		\node [style=none] (21) at (-0.75, -0.75) {};
		\node [style=none] (22) at (-1.75, 0.25) {};
		\node [style=none] (23) at (-1.75, 0) {};
		\node [style=none] (24) at (-1.75, -0.75) {};
		\node [style=none, scale=0.5] (25) at (-1, -0.25) {\vdots};
		\node [style=none, scale=0.5] (26) at (1.5, -0.25) {\vdots};
		\node [style=none, scale=0.5] (27) at (-2.25, -0.25) {m \Bigg\{};
		\node [style=none, scale=0.5] (28) at (2.75, -0.25) { \Bigg\} n};
		\node [style=none] (29) at (1, 0.25) {\scriptsize $\mathbf{t}$};
	\end{pgfonlayer}
	\begin{pgfonlayer}{edgelayer}
		\draw (16.center) to (12.center);
		\draw (17.center) to (13.center);
		\draw (18.center) to (14.center);
		\draw (22.center) to (19.center);
		\draw (23.center) to (20.center);
		\draw (24.center) to (21.center);
	\end{pgfonlayer}
\end{tikzpicture}}
    \end{center}

    \item \textit{Morphisms}. A morphism (\textbf{wiring diagram}) \(\Phi\) from a set of indexed boxes \(\mathbf{t}^{1},\ldots,\mathbf{t}^n\) to a box \(\mathbf{v}\) is a span in category \(\mathbf{Bij}_\tau\) which is the category of \(\tau\)-typed finite sets and bijective functions between them. The morphism is denoted as \(\Phi: \mathbf{t}^{1},\ldots,\mathbf{t}^{n} \rightarrow \mathbf{v}\). Pictorially \(\Phi\) is a wiring diagram with \(\mathbf{t}^{1},\ldots,\mathbf{t}^{n}\) as inner boxes and \(\mathbf{v}\) as outer box. More importantly, \(\Phi\) encodes how the wires are connected. Additionally, the indexed boxes \(\mathbf{t}^{1},\ldots,\mathbf{t}^{n}\) must satisfy a \emph{progress condition} which ensures that there is no cycle in a wiring diagram. For example:
    \begin{center}
        \scalebox{0.8}{\begin{tikzpicture}
	\begin{pgfonlayer}{nodelayer}
		\node [style=medium rounded box] (1) at (5.5, -0.25) {};
		\node [style=none] (3) at (2.25, -0.25) {};
		\node [style=none] (4) at (12, 0.25) {};
		\node [style=none] (5) at (6.5, -0.75) {};
		\node [style=none] (6) at (6.5, 0.25) {};
		\node [style=none] (7) at (7.75, -1.5) {};
		\node [style=medium rounded box] (8) at (8.75, -1.75) {};
		\node [style=none] (9) at (12, -1.75) {};
		\node [style=none] (10) at (2.25, -2.25) {};
		\node [style=none] (11) at (7.75, -2.25) {};
		\node [style=large rounded box] (12) at (7.25, -1) {};
		\node [style=none] (13) at (10, 0.75) {\textbf{v}};
		\node [style=none] (14) at (6, 0) {\scriptsize $\mathbf{t_1}$};
		\node [style=none] (15) at (9.25, -1.5) {\scriptsize $\mathbf{t_2}$};
	\end{pgfonlayer}
	\begin{pgfonlayer}{edgelayer}
		\draw (3.center) to (1);
		\draw (6.center) to (4.center);
		\draw [in=180, out=0, looseness=2.00] (5.center) to (7.center);
		\draw (9.center) to (8);
		\draw (11.center) to (10.center);
	\end{pgfonlayer}
\end{tikzpicture}}
    \end{center}
    \item \textit{Identities}. The identity morphism \(\mathbf{Id_t}\) on box \(\mathbf{t}\) is the identity span associated with \(\mathbf{t}\). Pictorially, it is a wiring diagram with no internal box, i.e., there is only one wire going right through the diagram. \vspace{-0.5cm}
    \begin{center}
        \scalebox{1.4}{\begin{tikzpicture}
	\begin{pgfonlayer}{nodelayer}
		\node [style=medium rounded box] (0) at (-0.25, -1) {};
		\node [style=none] (1) at (-1.75, -1) {};
		\node [style=none] (2) at (1.25, -1) {};
	\end{pgfonlayer}
	\begin{pgfonlayer}{edgelayer}
		\draw (1.center) to (2.center);
	\end{pgfonlayer}
\end{tikzpicture}
}
    \end{center}
    \item \textit{Composition formula}. Given wiring diagram \(\Psi: \mathbf{s}^{1}, \ldots, \mathbf{s}^{m} \rightarrow \mathbf{t}^{i} \text { and } \Phi: \mathbf{t}^{1}, \ldots, \mathbf{t}^{n} \rightarrow \mathbf{v}\), their \(i\)-th partial composition is denoted as \(\Psi {\fatsemi}_i \Phi\). We follow the notation from Patterson, Spivak and Vagner that \(\fatsemi\) gives composition of morphisms in operad and the order of composition is left to right. 
    The composition is a new wiring diagram:
    \[\Psi {\fatsemi}_i \Phi: \mathbf{t}^{1}, \ldots, \mathbf{t}^{i - 1}, \mathbf{s}^{1}, \ldots, \mathbf{s}^{m}, \mathbf{t}^{i + 1}, \ldots ,\mathbf{t}^{n} \rightarrow \mathbf{v}\]
    The formula says that the composition slots the whole wiring diagram \(\Psi\) into the inner box \(\mathbf{t}^{i}\) in wiring diagram \(\Phi\).
    \begin{center}
        \scalebox{0.7}{\begin{tikzpicture}
	\begin{pgfonlayer}{nodelayer}
		\node [style=large rounded box] (0) at (-6, 0) {};
		\node [style=medium rounded box] (1) at (-7.5, 0) {};
		\node [style=medium rounded box] (2) at (-4.25, 0) {};
		\node [style=none] (3) at (-10, 0) {};
		\node [style=none] (4) at (-3.25, 0.25) {};
		\node [style=none] (5) at (-3.25, -0.25) {};
		\node [style=none] (6) at (-2, 0.25) {};
		\node [style=none] (7) at (-2, -0.25) {};
		\node [style=none] (8) at (-0.75, 0) {$\fatsemi_i$};
		\node [style=medium rounded box] (9) at (3.75, 0.75) {};
		\node [style=none] (10) at (0.5, 0.75) {};
		\node [style=none] (11) at (10.25, 1.25) {};
		\node [style=none] (12) at (4.75, 0.25) {};
		\node [style=none] (13) at (4.75, 1.25) {};
		\node [style=none] (14) at (6, -0.5) {};
		\node [style=medium rounded box] (15) at (7, -0.75) {};
		\node [style=none] (16) at (10.25, -0.75) {};
		\node [style=none] (17) at (0.5, -1.25) {};
		\node [style=none] (18) at (6, -1.25) {};
		\node [style=large rounded box] (19) at (5.5, 0) {};
		\node [style=none] (20) at (11.25, 0) {=};
		\node [style=none] (23) at (22, 1) {};
		\node [style=none] (24) at (16.5, 0.5) {};
		\node [style=none] (25) at (16.5, 1) {};
		\node [style=none] (26) at (17.75, -0.5) {};
		\node [style=medium rounded box] (27) at (18.75, -0.75) {};
		\node [style=none] (28) at (22, -0.75) {};
		\node [style=none] (29) at (12.25, -1.25) {};
		\node [style=none] (30) at (17.75, -1.25) {};
		\node [style=large rounded box] (31) at (17.25, 0) {};
		\node [style=none] (35) at (12.25, 0.75) {};
		\node [style=small rounded box] (36) at (14.5, 0.75) {};
		\node [style=small rounded box] (37) at (16, 0.75) {};
		\node [style=medium dashed box] (38) at (15.25, 0.75) {};
	\end{pgfonlayer}
	\begin{pgfonlayer}{edgelayer}
		\draw (1) to (2);
		\draw (3.center) to (1);
		\draw (4.center) to (6.center);
		\draw (5.center) to (7.center);
		\draw (10.center) to (9);
		\draw (13.center) to (11.center);
		\draw [in=180, out=0, looseness=2.00] (12.center) to (14.center);
		\draw (16.center) to (15);
		\draw (18.center) to (17.center);
		\draw (25.center) to (23.center);
		\draw [in=180, out=0, looseness=2.00] (24.center) to (26.center);
		\draw (28.center) to (27);
		\draw (30.center) to (29.center);
		\draw (35.center) to (36);
		\draw (36) to (37);
	\end{pgfonlayer}
\end{tikzpicture}}
    \end{center}

    It can be shown that composition satisfies the progress condition which ensures that the wiring diagram is acyclic.
\end{itemize}
The necessary identity law and associativity law are left to readers to check.
\end{definition}

Operad algebras take composition structures in operads and map them to a meaning in some SMC. A $\mathcal{W}_{Ob(\mathcal{C})}$-algebra is an operad functor $H: \mathcal{W}_{Ob(\mathcal{C})} \rightarrow \mathcal{O}_{\mathcal{V}}$, where the latter is the operad underlying \(\mathcal{V}\).

\begin{proposition}[Enriched SMCs have enriched operad algebras]
\label{prop:enopalgebras}
Each \(\mathcal{V}\)-enriched strict SMC \(\mathcal{C}\) has an operad of wiring diagrams \(\mathcal{W}_{Ob(\mathcal{C})}\) and a \(\mathcal{V}\)-enriched algebra \(H: \mathcal{W}_{Ob(\mathcal{C})} \rightarrow \mathcal{O}_{\mathcal{V}}\). This operad provides a ``normal form'' for \(\mathcal{C}\), in the sense that the \(\mathcal{V}\)-enriched SMC arising from the algebra and \(\mathcal{C}\) itself.
\end{proposition}
\begin{proof}
Definition 6.63 in Fong and Spivak~\cite{Fong2019} shows how to construct the operad \(\mathcal{O}_{\mathcal{V}}\) underlying \(\mathcal{V}\). Theorem 5.3 in Patterson~\cite{Patterson2021} then shows how to obtain the algebra \(H: \mathcal{W}_{Ob(\mathcal{C})} \rightarrow \mathcal{O}_{\mathcal{V}}\), substituting \(\mathcal{O}_\mathcal{V}\) for the operad \texttt{Set} and hom-objects for hom-sets.
\end{proof}
We therefore understand the \(\mathcal{V}\)-enriched algebra $H$ as
\begin{itemize}
    \item Sending a box $(\mathbf{t}_{-}, \mathbf{t}_{+}) \in \mathcal{W}_{Ob(\mathcal{C})}$ to the hom-object $\mathcal{C}(\mathbf{t}_{-}, \mathbf{t}_{+}) \in \mathcal{V}$ whose shapes it specifies.
    \item Sending a wiring diagram $\Phi: \mathbf{t}^1 \times \ldots \times \mathbf{t}^n \rightarrow \mathbf{v} $ to the morphism $H(\Phi): \mathcal{C}(\mathbf{t}^{1}_{-}, \mathbf{t}^{1}_{+}) \otimes \ldots \otimes \mathcal{C}(\mathbf{t}^{n}_{-}, \mathbf{t}^{n}_{+}) \rightarrow \mathcal{C}(\mathbf{v}_{-}, \mathbf{v}_{+})$ from the boxes' hom-objects to the whole diagram's hom-object.
\end{itemize}

\vspace{-1em}
\subsection{Free monoidal categories over quivers}
\label{subsec:freecats}
Learning structures of composition will require specifying, a priori, some set of generating objects and morphisms between them. However, a specific structure of possible compositions may have semantics in several concrete categories, depending on the needs of the application. The free category prior therefore defers semantics to specific applications, considering only structure in the generative model itself.

The free category prior requires, as a hyperparameter, a directed multigraph $G=(V, E)$ as a nerve. Each vertex $v \in V$ must have a label $\sigma(v)$ drawn from $\bigcup_{n \in \mathbb{N}} \Sigma^{n}$ (the Kleene closure of a finite symbol-set $\Sigma$), and each label in $\Sigma^1$ must appear on a vertex. A unique label $I \in \Sigma$ is reserved to denote in the graph the vertex underlying the monoidal unit. Each edge has a ``domain'' $\text{dom}(e)$ and a ``codomain'' $\text{cod}(e)$. We write $E(A, B)$ to denote the edge-set between $A, B \in V$, and write (assuming \(\emptyset = E^0 \))
\begin{align*}
    P(A, B) &:= \bigcup_{n\in\mathbb{N}} \{p\in E^{n} : \text{dom}(p_1) = A, \text{cod}(p_n) = B \}
\end{align*}
for the path-set from $A$ to $B$. We impose the following condition on the nerve-graph $G$.

\begin{condition}[All non-unit generating objects have points]
\label{cond:reachability}
Each vertex $v\in V$  with $|\sigma(v)|=1$, $\sigma(v)\neq I$, and a nonzero out-degree must have a path into it from the unit
\begin{align*}
    \forall v\in V. (|\sigma(v)| = 1) \wedge (\text{deg}^{+}(v) > 0) \rightarrow (|P(I, v)| > 0).
\end{align*}
\end{condition}

\begin{definition}[Directed multigraph with monoidal points]
\label{def:monoidalpoints}
Given a directed multigraph $G$ meeting Condition~\ref{cond:reachability}, the quiver $Q = (V_Q, E_Q)$ where
\begin{align*}
    V_Q &= V \cup \{I\} \\
    E_Q &= E \cup \bigcup_{ \{ (\sigma(v_1) \sigma(v_2)) \in V : (I, v_1), (I, v_2) \in E \}} \{(I, \sigma(v_1) \sigma(v_2))\}
\end{align*}
has paths from the unit vertex $I$ to its vertices with labels longer than one symbol. Adding the extra edges to the quiver ensures that any two generating vertices with paths from the unit give rise to a compound vertex with a path from the unit. Note that the original multigraph $G$ may already include a vertex for $I$; this construction simply ensures one is added if it doesn't.
\end{definition}

\begin{definition}[Free monoidal category on a directed multigraph]
\label{def:freecat}
Assume that $Q=(V_Q, E_Q)$ is a quiver constructed via Definition~\ref{def:monoidalpoints}. The free monoidal category $\freeCat{Q}$ (notation due to Gavranovic~\cite{Gavranovic2019}) on that quiver has as objects finite Cartesian products of vertex labels
\begin{align*}
    Ob(\freeCat{Q}) &:= \bigcup_{n \in \mathbb{N}} \{\sigma(v) : v \in V_Q\}^{n}.
\end{align*}
We define the hom-sets for the free monoidal category inductively. For each
\(A, B \in V_Q\) such that \(|\sigma(A)|=1, |\sigma(B)|=1\),
\begin{align*}
  \freeCat{Q}(\sigma(A), \sigma(B)) := P(A, B).  
\end{align*}
For $A, B, C, D \in \freeCat{Q}$, none of which equals $I$,
\begin{align*}
    \freeCat{Q}(\sigma(A) \sigma(C), \sigma(B) \sigma(D)) &:= \{(f, g) : f \in \freeCat{Q}(\sigma(A), \sigma(B)), g \in \freeCat{Q}(\sigma(C), \sigma(D)) \}.
\end{align*}
Finally, for each \(A \in \freeCat{Q}\) we define the identity morphism to be
\begin{align*}
    id_A &= \emptyset \in \freeCat{Q}(A, A).
\end{align*}
The above construction gives the free (Cartesian) monoidal product on paths in a graph, with the monoidal action on objects being concatenation of vertex labels. This category has as morphisms finite tuples of paths between the respect elements of finite tuples of vertices.
\end{definition}

\vspace{-1em}
\section{Probabilistic categorical generative modeling}
\label{sec:categoricalpgm}

\subsection{Assigning probabilities to morphisms by enrichment in a Markov category}
\label{subsec:homdists}
Definition~\ref{def:freecat} defined the free category over a given quiver, giving a purely syntactic description of compositional structure.
Constructing a probabilistic generative model over such structure will require defining a version of the free category $\freeCat{Q}$ enriched in a Markov category, in our case \textbf{QBS}. This requires first showing that hom-sets in $\freeCat{Q}$ admit the structure of objects in the enriching category.
\begin{lemma}[Hom-sets in free (monoidal) categories admit the structure of quasi-Borel spaces]
\label{lem:randompaths}
Let $G$ be a directed multigraph and $\freeCat{Q}$ its free monoidal category as defined above. Each hom-set $\freeCat{Q}(A, B)$ can admit the structure of a quasi-Borel space.
\end{lemma}
\begin{proof}
An object in \textbf{QBS} is a pair $(X, M_X)$ with $X$ a set and $M_X \subseteq [\mathbb{R} \rightarrow X]$ a set of primitive random variables over that space. For each $A, B \in V$ we already have a hom-set $\freeCat{Q}(A, B)$, and just need to construct a corresponding set of random variables. We take advantage of the countable-discrete nature of the underlying path-set $\freeCat{Q}(A, B)$ to construct the standard discrete $\sigma$-algebra over the set, which we call $\Sigma_{Q(A, B)}$. We then call $M_{B^A}$ the set of measurable functions $\mathbb{R} \rightarrow \freeCat{Q}(A, B)$. We therefore have $(\freeCat{Q}(A, B), M_{B^A})) \in \textbf{QBS}$ as required.
\end{proof}

Defining quasi-Borel spaces with path-sets as their sample spaces enables defining a probabilistically enriched free category. Recall from Lemma~\ref{lem:probMonad}, $\prob{}$ is the probability monad in \textbf{QBS}. Furthermore, we define the enrichment here with a twist: since we require a probabilistic generative model, composition builds up joint distributions, without marginalizing away intermediate spaces.
\begin{definition}[Probabilistic generative free category]
\label{def:pgcat}
The probabilistic generative category $\p{Q}$ is a version of $\freeCat{Q}$ enriched in $\mathbf{QBS}$. We construct it by specifying
\begin{itemize}
    \item Its objects $Ob(\p{Q}) = Ob(\freeCat{Q})$;
    \item For each $(A, B) \in Ob(\p{Q}) \times Ob(\p{Q})$ the hom-object \(\p{Q}(A, B) = \prob{(\freeCat{Q}(A, B), M_{B^A})} \in \mathbf{QBS}\);
    \item For each $A \in Ob(\p{Q})$ the identity element
    \begin{align*}
        id_{A} &: \mathbf{QBS}(I, \prob{(\freeCat{Q}(A, A), M_{A^A})}) \\
        id_{A} &= I \mapsto \eta_{\freeCat{Q}(A, A)}(\mathbf{const}_{\emptyset}: (\freeCat{Q}(A, A), M_{A^A}))
    \end{align*}
    gives a Dirac delta distribution over the empty path in \(\freeCat{Q}(A, A)\);
    \item For each $A, B, C \in Ob(\p{Q})$ a way of composing morphisms, in this case
    \begin{align*}
        \fatsemi_{A,B,C} &: \p{Q}(A, B) \odot \p{Q}(B, C) \rightarrow \p{Q}(A, C) \\
        f \fatsemi_{A,B,C} g &= (f \odot_{\mathbf{QBS}} g) \fatsemi_{\prob{}} (x \odot y \mapsto \eta_{\freeCat{Q}(A, C)}(x \fatsemi_{\freeCat{Q}} y)),
    \end{align*}
    pushes-forward the joint distribution defined by $f$ and $g$ into a distribution over $\freeCat{Q}(A, C)$.
    \item \(\p{Q}\) inherits its monoidal structure from \textbf{QBS}
    \begin{align*}
        \odot_{A, B, C D} &: \p{Q}(A, B) \times \p{Q}(C, D) \rightarrow \p{Q}(A \odot C, B \odot D) \\
        f \odot_{A, B, C, D} g &= (f \odot_{\mathbf{QBS}} g).
    \end{align*}
    The usual probabilistic composition uses a random outcome from one distribution to parameterize another. Here we take their monoidal product, representing their joint distribution.
\end{itemize}
\end{definition}
Above we defined operads $\mathcal{W}_{Ob(\mathcal{C})}$ of wiring diagrams, and algebras over them. Now, having an enriched category $\p{Q}$, we can demonstrate the existence of an enriched operad algebra mapping wiring diagrams $\Phi$ into $\p{Q}$.
\begin{lemma}[Existence of algebras of wiring diagrams over the probabilistic generative free category]
\label{lem:algebras}
We consider the concrete setting where $\mathcal{C}=\p{Q}$, the probabilistic generative free category, with the enriching category being $\mathcal{V}=\mathbf{QBS}$. An operad algebra in this setting has the form
\(H: \mathcal{W}_{Ob(\mathcal{C})} \rightarrow Op(\mathbf{QBS})\), which sends a box $\mathbf{t} = (\mathbf{t}_{-}, \mathbf{t}_{+}) \in \mathcal{W}_{Ob(\mathcal{C})}$ to the quasi-Borel space $\p{Q}(\mathbf{t}_{-}, \mathbf{t}_{+}) \in \mathbf{QBS}$ and the collection of hom-distributions in a wiring diagram $\Phi$ to their joint distribution within the space of distributions over paths from the input and output types of the diagram as a whole.
\end{lemma}
\begin{proof}
Definition~\ref{def:pgcat} (of $\p{Q}$) gives the machinery for forming $\tau$-typed hom-distributions, and joint hom-distributions, in \textbf{QBS}. Since \textbf{QBS} is an SMC and $\p{Q}$ inherits its monoidal structure, $\p{Q}$ supports an operad algebra as an instantiation of Proposition~\ref{prop:enopalgebras} above. This construction requires that the hom-distribution assigned to each box in a wiring diagram be conditionally independent of all others, given the wiring diagram itself.
\vspace{-1em}
\end{proof}

\vspace{-1em}
\subsection{The free category prior over morphisms}
\label{subsec:freecatprior}
Definition~\ref{def:pgcat} demonstrates that free categories \emph{allow} enrichment with joint distributions, and similarly, Lemma~\ref{lem:algebras} demonstrates the existence of operad algebras taking wiring diagrams into joint distributions over morphisms in the free category. These existence proofs alone do not constitute a generative model: that requires a distribution with constructive procedures for sampling and density evaluation. This section will define such procedures and the joint distribution they induce on morphisms.

A probabilistic generative model over some space must generally impose some notion of simplicity, assigning higher prior probability to simpler elements of the space. Free monoidal categories have only two kinds of structure that give rise to complexity: sequential and parallel composition. We hypothesize that path-length and product-width in the underlying quiver may serve as a reasonable inductive bias. Shorter, ``narrower'' paths should have higher probability, but no path should have zero probability.

We thus begin with our quiver $Q$ representing the free category, and a wiring diagram $\Phi$. Each box $\mathbf{t}^{i} \in \text{dom}(\Phi)$ consists of a source and target $\mathbf{t}^{i}_{-}, \mathbf{t}^{i}_{+} \in V_Q$. To generate morphisms, we need a policy
\begin{align*}
    \pi(e \mid v, \mathbf{t}_{+}) &: E_Q(v, \cdot) \rightarrow (0, 1]
\end{align*}
that randomly selects edges out of $v$, yielding short paths to the destination $\mathbf{t}_{+}$. This is a stochastic short-paths problem, generalizing the all-destinations shortest-path problem (which has itself inspired a learning problem \cite{Jurgenson2020}) to a setting of probabilistic transitions.

We construct our stochastic short-paths problem by transforming the original nerve quiver $Q$ into a simple directed bipartite graph $G'=(V', E')$ for
\begin{align*}
   V' &= V_Q \cup E_Q
   &
   E' &= \bigcup_{e \in E_Q} \{ (\text{dom}(e), e), (e, \text{cod}(e))  \}.
\end{align*}
This directed graph has two sets of vertices: one of the generating objects $v \in V_Q$ and another of the generating morphisms $e \in E_Q$. In this graph, generating object vertices only receive edges from generating morphism vertices, and vice-versa, while each generating morphism vertex has only a single exiting edge. No two vertices have multiple edges between them.

\begin{lemma}[Equivalence of free categories under conversion from quiver to directed bipartite graph]
\label{lem:graphequivalence}
The original free category $\freeCat{Q}$ is equivalent to the one generated by taking the directed bipartite graph $G'$, converting it back into a quiver $Q'$ according to the above construction, and generating a free category $\freeCat{Q'}$ from that.
\end{lemma}
\begin{proof}
Representing both graphs via their adjacency matrices will demonstrate equivalence of the graphs, with proofs due to Brualdi~\cite{Brualdi1980} and others as old results. Equivalence of the underlying graphs then produces an equivalence of their free categories.
\end{proof}
We represent $Q$ as $G'$ because the latter supports representing our policy as a transition kernel $\pi \in \mathbf{QBS}(V', V')$. Given the directed adjacency matrix $A' \in \mathbb{R}^{|V'| \times |V'|}$, Estrada and Hatano~\cite{Estrada2008} defined the communicability between $\mathbf{t}^{i}_{-}$ and $\mathbf{t}^{i}_{+}$ as
\begin{align}
    C_{\mathbf{t}^{i}_{-},\mathbf{t}^{i}_{+}} &= \sum_{n=1}^{\infty} \frac{({A'}^{l})_{\mathbf{t}^{i}_{-}, \mathbf{t}^{i}_{+}}}{n!} = (e^{A'})_{\mathbf{t}^{i}_{-}, \mathbf{t}^{i}_{+}}.
\end{align}
This communicability measure, also known as the matrix exponential, measures the number of paths in the underlying graph, weighted by their length. Its negative logarithm
\begin{align}
\label{eq:quasimetric}
    d(\mathbf{t}^{i}_{-}, \mathbf{t}^{i}_{+}) &= -\log (C)_{\mathbf{t}^{i}_{-}, \mathbf{t}^{i}_{+}}   
\end{align}
provides a quasimetric\footnote{Lacking symmetry due to directedness of edges.} of ``intuitive distance'' on the graph \cite{baram2018intuitive}. Intuitive distance in a graph penalizes actual path length, while rewarding paths connecting many sources and targets.

We therefore parameterize the policy in log-space $\log \pi(e \mid v; \mathbf{t}_{+}, Q)$, for global hyperparameters $Q$ and $\mathbf{t}_{+} \in V_Q$, with global random variables
\begin{itemize}
    \item $\mathbf{w} \sim \mathcal{N}(\Vec{0}, \Vec{1}) \in \mathbb{R}^{|V'|}$: a vector of single-step ``preferences'' over objects and morphisms sampled from a standard Gaussian prior;
    \item $\beta \sim \gamma(1, 1) \in \mathbb{R}_{+}$: a positive inverse-temperature specifying the ``confidence'' of the policy, sampled from a Gamma prior;
\end{itemize}
and the local random variable $v \in V_Q$. $v$ is a ``present location'' in the graph $G'$, obtained autoregressively. The policy is then written in terms of its surprisal as
\begin{align}
\label{eq:policy}
    -\log \pi(e \mid v, \mathbf{w}, \beta; \mathbf{t}_{+}, Q) &\propto -\frac{1}{\beta} \log \left(C + (A')\text{diag}(\mathbf{w}) \right)_{\text{cod}(e), \mathbf{t}_{+}},
\end{align}
and we have the following theorem relating the policy to the global structure of the quiver $Q$.
\begin{theorem}[The intuitive distance lower-bounds expected policy surprisal]
\label{thm:distancebound}
Under the prior distributions on the global random variables, the expected unnormalized surprisal of the policy (Equation~\ref{eq:policy}) is at least the intuitive distance (Equation~\ref{eq:quasimetric})
\begin{align*}
    d(\text{cod}(e), \mathbf{t}_{+}) &\leq \expect{
        \mathbf{w} \sim \mathcal{N},
        \beta \sim \gamma(1, 1)
    }{
        -\log \pi(e \mid v, \mathbf{w}, \beta; \mathbf{t}_{+}, Q)
    }.
\end{align*}
\end{theorem}
\begin{proof}
We begin by expanding the definition of the expectation
\begin{align*}
    \expect{
        \mathbf{w} \sim \mathcal{N},
        \beta \sim \gamma(1, 1)
    }{
        -\log \pi(e \mid v, \mathbf{w}, \beta; \mathbf{t}_{+}, Q)
    }
    &=
    \expect{\mathbf{w} \sim \mathcal{N}}{
        \expect{
            \beta \sim \gamma(1, 1)
        }{
            -\frac{1}{\beta} \log \left(C + (A')\text{diag}(\mathbf{w}) \right)_{\text{cod}(e), \mathbf{t}_{+}}
        }
    }
\end{align*}
and recognizing that our Gamma prior has a mean of 1. Substituting the inner expectation away and cancelling, we have
\begin{align*}
    \expect{
        \mathbf{w} \sim \mathcal{N},
        \beta \sim \gamma(1, 1)
    }{
        \log \pi(e \mid v, \mathbf{w}, \beta; \mathbf{t}_{+}, Q)
    }
    &=
    \expect{\mathbf{w} \sim \mathcal{N}}{
        \log \left(C + (A')\text{diag}(\mathbf{w}) \right)_{\text{cod}(e), \mathbf{t}_{+}}
    }.
\end{align*}
Jensen's inequality for log-expectations says an expected-log lower bounds a log-expectation
\begin{align*}
    \expect{
        \mathbf{w} \sim \mathcal{N},
        \beta \sim \gamma(1, 1)
    }{
        \log \pi(e \mid v, \mathbf{w}, \beta; \mathbf{t}_{+}, Q)
    }
    &\leq
    \log \left(\expect{\mathbf{w} \sim \mathcal{N}}{
        C + (A')\text{diag}(\mathbf{w})
    } \right)_{\text{cod}(e), \mathbf{t}_{+}} \\
    &\leq 
    \log \left(
        C + \expect{\mathbf{w} \sim \mathcal{N}}{(A')\text{diag}(\mathbf{w})}
    \right)_{\text{cod}(e), \mathbf{t}_{+}},
\end{align*}
and we can now substitute in the standard normal's mean
\begin{align*}
    \expect{
        \mathbf{w} \sim \mathcal{N},
        \beta \sim \gamma(1, 1)
    }{
        \log \pi(e \mid v, \mathbf{w}, \beta; \mathbf{t}_{+}, Q)
    }
    &\leq
    \log \left(
        C + (A')\text{diag}(\Vec{0})
    \right)_{\text{cod}(e), \mathbf{t}_{+}} \\
    &\leq
    \log \left(
        C
    \right)_{\text{cod}(e), \mathbf{t}_{+}} \\
    -\log \left(
        C
    \right)_{\text{cod}(e), \mathbf{t}_{+}}
    &\leq
    \expect{
        \mathbf{w} \sim \mathcal{N},
        \beta \sim \gamma(1, 1)
    }{
        -\log \pi(e \mid v, \mathbf{w}, \beta; \mathbf{t}_{+}, Q)
    } \\
    d(\text{cod}(e), \mathbf{t}_{+})
    &\leq
    \expect{
        \mathbf{w} \sim \mathcal{N},
        \beta \sim \gamma(1, 1)
    }{
        -\log \pi(e \mid v, \mathbf{w}, \beta; \mathbf{t}_{+}, Q)
    }.
\end{align*}
\vspace{-1em}
\end{proof}
The policy's parametric form thus normalizes to
\begin{align}
\label{eq:normedpolicy}
    \pi(e \mid v, \mathbf{w}, \beta; \mathbf{t}_{+}, Q) &= \text{softmin}\left(
        -\frac{1}{\beta} \log(C + (A')\text{diag}(\mathbf{w}))_{\cdot, \mathbf{t}_{+}}
    \right)_{\text{cod}(e)}.
\end{align}
Theorem~\ref{thm:distancebound} demonstrates that Equation~\ref{eq:normedpolicy} assigns ``energy'' to finite edge-sets proportionately to a stochastic upper bound on their intuitive distance, scales the energies according to its ``confidence'' $\beta$, and employs soft minimization to assign probabilities to edges. When arrival to $v=\mathbf{t}_{+}$ samples a path $f= (e_1, \ldots, e_L): \mathbf{t}_{-} \rightarrow \mathbf{t}_{+}$ of length $L$, it has probability
\begin{align}
\label{eq:pmorphism}
    \mathbb{P}(f \mid \mathbf{w}, \beta; \mathbf{t}_{-}, \mathbf{t}_{+}, Q) &= \pi(e_1 \mid \mathbf{t}_{-}, \mathbf{w}, \beta; \mathbf{t}_{+}, Q) \prod_{l=2}^{L} \pi(e_l \mid \text{cod}(e_{l-1}), \mathbf{w}, \beta; \mathbf{t}_{+}, Q).
\vspace{-1em}
\end{align}
Equation~\ref{eq:pmorphism} relies on the explicit quiver representation $Q$, and so can only ``connect'' generator objects $\mathbf{t}_{-}, \mathbf{t}_{+} \in V_Q$ to define a hom-distribution $\p{Q}(\mathbf{t}_{-}, \mathbf{t}_{+})$. However, the second term defining $E_Q$ in Definition~\ref{def:monoidalpoints} defines a series of edges in $Q$ mapping $I \rightarrow \p{Q}(\mathbf{t}^{1}_{+} \odot \mathbf{t}^{2}_{+})$. These edges represent ``points'' in the product object, and their semantic content is to recursively invoke the free category prior to generate \(f: I \rightarrow \mathbf{t}^{1}_{+}\), \(g: I \rightarrow \mathbf{t}^{2}_{+}\), and then finally combine them into \(f \odot g: I \rightarrow \mathbf{t}^{1}_{+} \odot \mathbf{t}^{2}_{+}\).

We employ ``macro expansion'' rather than split and simplify boxes in wiring diagrams because some applications may supply their own generating morphisms into product objects. Our approach here simply supplies points in product objects when the component objects support points.

\vspace{-1em}
\subsection{The complete generative model over wiring diagrams}
\label{subsec:joint}
In summary, given a quiver $Q$ and a wiring diagram $\Phi: \mathbf{t}^{1} \times \ldots \times \mathbf{t}^{n} \rightarrow \mathbf{v}$, the sampling process for generating a morphism $f \sim \p{Q}$ fitting $\Phi$ proceeds as
\begin{align*}
    \beta &\sim \gamma(1, 1) \\
    \mathbf{w} &\sim \mathcal{N}(0, 1) \\
    \forall i \in [1..n]. f_{i} &\sim \mathbb{P}(\mathbf{f}_{i} \mid \mathbf{w}, \beta; \mathbf{t}^{i}_{-}, \mathbf{t}^{i}_{+}, Q).
\end{align*}
This procedure induces a joint distribution over all latent variables
\begin{align}
    \label{eq:generative}
    p(f, \mathbf{w}, \beta; \Phi, Q) &= p(\beta) p(\mathbf{w}) \prod_{\textbf{t} \in \Phi} \mathbb{P}(f_{\textbf{t}} \mid \mathbf{w}, \beta; \textbf{t}, Q).
\end{align}
Finally, in order to learn morphisms, an application must supply both a concrete semantic functor $F: \freeCat{Q} \rightarrow \mathcal{C}$ into an arbitrary SMC $\mathcal{C}$ and a likelihood $p(x \mid F(f))$ relating morphisms to data.
\begin{theorem}[Bayesian learning with free category priors]
Assuming that the semantic category $\mathcal{C}$ supports enrichment in \textbf{QBS} via joint distributions, the free category prior indeed acts as a prior. Sampling a morphism $f \sim p(f; \Phi, Q)$ from the prior, assigning semantics $F(f)$ to that morphism, and evaluating a likelihood $p(x \mid F(f))$ relating those semantics to data induces a joint density and Bayesian inverse.
\end{theorem}
\begin{proof}
We can ``lift'' the semantics $F: \freeCat{Q} \rightarrow \mathcal{C}$ into a functor $\prob{F}: \p{Q} \rightarrow \mathbf{QBS}-\mathcal{C}$ by defining the action on morphisms
\begin{align*}
    \prob{F}(\cdot) &: \p{Q}(A, B) \rightarrow \mathbf{QBS}-\mathcal{C}(F(A), F(B)) \\
    \prob{F}(p) &= p \fatsemi_{\prob{}} (f \mapsto \eta_{F(B)}(F(f))),
\end{align*}
to push-forward \(\p{Q}(A, B)\) via $F(f)$. The semantics and likelihood then define a joint density
\begin{align}
    \label{eq:genjoint}
    p(x, f, \mathbf{w}, \beta; \Phi, Q) &= p(x \mid F(f)) p(f, \mathbf{w}, \beta; \Phi, Q)
\end{align}
with a Bayesian inversion
\begin{align}
\label{eq:bayesinverse}
    p(f, \mathbf{w}, \beta \mid x; \Phi, Q) &= \frac{
        p(x, f, \mathbf{w}, \beta; \Phi, Q)
    }{
        p(x; \Phi, Q)
    }.
\end{align}
\end{proof}
Section~\ref{sec:software} will discuss the software implementation of the free category prior, and an experiment in which a concrete quiver $Q$ is assigned specific semantics and likelihood to perform learning.

\vspace{-1em}
\section{Machine learning with the free category prior}
\label{sec:software}

For machine learning applications, we provide a software implementation of the free category prior. We call it \href{https://github.com/neu-pml/discopyro}{Discopyro}; it is available as a Python library on top of the Discopy~\cite{DeFelice2020} library for computing with morphisms in monoidal categories and the Pyro~\cite{bingham2019pyro} probabilistic programming language. The Discopyro implementation takes the directed multigraph $G$ as a hyperparameter; constructs its corresponding $Q$, $\freeCat{Q}$, and $\p{Q}$; and uses those to conditionally sample morphisms for application-specific wiring diagrams $\Phi$. The Discopyro software supports implicitly associating each edge (generating morphism) $e \in E_Q$ with an instance of \texttt{discopy.monoidal.Diagram}, representing an arbitrary SMC \(\mathcal{C}\) via Discopy.  The user can also apply a Discopy \texttt{Functor} defining the semantics \(F: \freeCat{Q} \rightarrow \mathcal{C}\) to interpret morphisms into the chosen category.

\vspace{-1em}
\subsection{Implementing the sampling process}
\label{subsec:sampling}

To specify wiring diagrams in Python, we have added wiring diagrams to a \href{https://github.com/neu-pml/discopy}{fork of Discopy}. Our implementation is fairly naive and just consists of \texttt{Box}, \texttt{Id} wires, and \texttt{Sequential} and \texttt{Parallel} composites, all subclasses of a \texttt{wiring.Diagram} base class. These come equipped with a \texttt{collapse()} method which implements the canonical catamorphism on the algebraic data type of wiring diagrams. We then implement wiring-diagram algebras for sampling from the $\p{Q}$ representation via F-algebras on wiring diagrams. Python pseudocode describing Discopyro's implementation will denote sequential (categorical) and parallel (monoidal product) compositions by their Discopy operators \texttt{>>} and \texttt{@}.

We represent the free category prior defined in Section~\ref{subsec:freecatprior} as a program that samples from the model in the Pyro~\cite{bingham2019pyro} probabilistic programming language. We give the name \texttt{path\_through} to the sample procedure defined above, and give pseudocode for its implementation in Listing~\ref{alg:path-between}.

\begin{listing*}[h]
    \begin{minted}[escapeinside=||,mathescape=true]{python}
    def path_through(self, wbox, energies, temperature):
        loc = wbox.dom
        f = |$id_A$|
        while loc != wbox.cod:
            logits = [energies[a, wbox.cod] for a in self.out_arrows(loc)]
            idx = pyro.sample(Categorical(logits=logits / temperature))
            arrow = self.out_arrows(loc)[idx]
            if not isinstance(arrow, Box):
                wiring = [wiring.Box(|$I$|, ob)] for ob in arrow.dest.objects]
                wiring = reduce(|$\lambda$| x, y: x @ y, wiring, wiring.Id(|$I$|))
                arrow = self.sample_morphism(wiring, energies)
            f = f >> arrow # Composition of morphisms
            loc = arrow.cod
        return f
    \end{minted}
    \caption{Generative model for short paths from $A$ to $B$ in the nerve $Q$.}
    \label{alg:path-between}
    \vspace{-1em}
\end{listing*}

Listing~\ref{alg:sample-morphism} then extends the short-paths procedure from individual boxes to whole wiring diagrams, implementing the operad algebra of Lemma~\ref{lem:algebras}. Since probability distributions are once again represented by random samplers and wiring diagrams by combinatorial data structures, the algebra over wiring diagrams is written as an $F$-algebra over the wiring diagrams themselves.

\begin{listing*}[h]
    \begin{minted}[escapeinside=||,mathescape=true]{python}
    def __sampler_falg__(self, f, energies, temp):
        if isinstance(f, Id):
            return ar_factory.id(f.dom)
        if isinstance(f, Box):
            return self.path_through(f, energies, temp)
        if isinstance(f, Sequential):
            return reduce(|$\lambda$| f, g: f >> g, arrows)
        if isinstance(f, Parallel):
            return reduce(|$\lambda$| x, y: x @ y, factors, ar_factory.id(Ty()))
    
    def sample_morphism(self, wdiagram, energies, temperature):
        falg = |$\lambda$| f: self.__sampler_falg__(f, energies, temperature)
        return wdiagram.collapse(falg)
    \end{minted}
    \caption{The operad functor mapping wiring diagrams to morphisms in $\freeCat{Q}$}
    \label{alg:sample-morphism}
    \vspace{-1em}
\end{listing*}

\vspace{-1em}
\subsection{Amortized inference with the free category prior}
\label{subsec:amortizedinf}
Learning morphisms from data via a likelihood requires approximately estimating the marginal likelihood $p(x; \Phi, Q)$ and thereby approximating the Bayesian inversion (Equation~\ref{eq:bayesinverse}). Discopyro provides amortized variational inference over its own random variables via neural proposals, whose parameters we label $\phi$, for the ``confidence'' $\beta \sim q_\phi(\beta \mid x)$ and the edge ``preferences'' $\mathbf{w} \sim q_\phi(\mathbf{w} \mid x)$.

The samples from these proposal distributions then parameterize a proposal, identical to the generative model, over $f_\theta \sim \mathbb{P}(f_\theta \mid \mathbf{w}, \beta; \Phi, Q)$. This gives us a complete proposal for the latent variables induced by the free category prior itself, 
\begin{align*}
    q_\phi(f_\theta, \mathbf{w}, \beta \mid x; \Phi, Q) &= \mathbb{P}(f_\theta \mid \mathbf{w}, \beta; \Phi, Q) q_\phi(\beta \mid x) q_\phi(\mathbf{w} \mid x).
\end{align*}

Finally, if an application programmer wants to make use of Discopyro's amortized inference functionality, they must supply a proposal $q_\phi(F(f_\theta) \mid f_\theta, x)$ which approximates the posterior distribution over the morphism's semantics $F(f_\theta)$. This final piece induces a joint proposal density
\begin{align}
    \label{eq:jointproposal}
    q_\phi(F(f_\theta), f_\theta, \mathbf{w}, \beta \mid x; \Phi, Q) &= q_\phi(F(f_\theta) \mid f_\theta, x) q_\phi(f_\theta, \mathbf{w}, \beta \mid x; \Phi, Q),
\end{align}
on the parameters $\mathbf{w}$ and $\beta$, the string diagram structure $f_\theta$ (with internal parameters $\theta$) compatible with the wiring diagram $\Phi$, and the semantics $F(f)$. Maximizing the ELBO (Equation~\ref{eq:elbo})
\begin{align*}
    \mathcal{L}(\theta, \phi) &= \expect{q_\phi}{\log \frac{p_\theta(x, F(f_\theta), f_\theta, \mathbf{w}, \beta; \Phi, G)}{q_\phi(F(f_\theta), f_\theta, \mathbf{w}, \beta \mid x)}}
\end{align*}
by stochastic gradient ascent adjusts the proposal to approximate the true posterior distribution (a process called variational Bayesian inference \cite{Blei2016}). Appendices \ref{app:importanceweighting} and \ref{app:elbo} derive and justify this objective function.

\subsection{Experiment: representing Omniglot with deep generative models}
\label{subsec:experiments}

\begin{table}[t]
    \centering
    \begin{tabular}{c|c|c|c|c}
        \toprule
        Model & Image Size & Learns Structure & log-$\Hat{Z}$ & log-$\Hat{Z}$/dim \\
        \hline 
        Sequential Attention & 28x28 & \xmark & -95.5 & -0.1218 \\
        Variational Homoencoder (PixelCNN) & 28x28 & \xmark & -61.2 & -0.0780 \\
        Graph VAE & 28x28 & \cmark & -104.6 & -0.1334 \\
        Generative Neurosymbolic & 105x105 & \cmark & -383.2 & -0.0348 \\
        Free Category DGM (ours) & 28x28 & \cmark & -11.6 & \textbf{-0.0148} \\
        \bottomrule
    \end{tabular}
    \caption{Average log model evidence on the Omniglot evaluation set across deep generative models}
    \label{tab:loglikelihoods}
    \vspace{-1em}
\end{table}

As a demonstrative experiment, we constructed a graphical nerve $G$ whose edges (generating morphisms) implemented probabilistic programs in Pyro by inheriting from \texttt{discopy.cartesian.Box}. We trained the resulting free category model on the Omniglot challenge dataset for few-shot learning \cite{Lake2019b}.

Internally, the edges implemented functions consisting of neural building blocks (ie: neural networks elements with probabilistic sampling) for deep generate models. Each generating morphism consisted not only of a domain, codomain, and function but also of a trace-type for its random sampling. Lew et al.~\cite{Lew2020} gave semantics in \textbf{QBS} to traced probabilistic programs  with a monoidal structure over traces\footnote{Note that $\tau$ here refers to trace types, not to wire types in a wiring diagram.}
\begin{align*}
    \fatsemi_{\prob{}_{\tau}} &: \mathbf{QBS}(A, \prob{(B \times \tau)}) \times \mathbf{QBS}(B, \prob{(C \times \tau)}) \rightarrow \mathbf{QBS}(A, \prob{(C \times \tau)}).
\end{align*}
This additional monoidal structure induces an appropriately ``traced''\footnote{In the sense of program execution, not traced categories.} probability monad and Kleisli category, which we call $\prob{\mathbf{QBS}}_{\tau}$ as a subcategory of \textbf{QBS}. The implied semantics functor and likelihood function apply the monadic multiplier of $\mathbb{P}$ in \textbf{QBS} to yield only a single-level probability endofunctor. Our experiment assumes data $x \in \mathbb{R}^{28 \times 28}$, and that morphisms therefore induce the joint likelihood
\begin{align*}
    p_\theta(x \mid z, f) &= \mathcal{N}(\mu{_\theta}(z, f), \tau I) \\
    p(x \mid F(f)) &= p_\theta(x \mid z, f) p_\theta(z \mid f)  
\end{align*}
and semantics proposal over the random variable $x$
\begin{align*}
    q_\phi(F(f) \mid f, x) &= q_\phi(z \mid f^{\dagger}_{\phi}, x).
\end{align*}

The random variable $z$ denotes unobserved random variables occurring in the randomness trace when running $f_\theta$ on data $x$. $f^{\dagger}_{\phi}$ is the result of an endofunctor $F_q: \prob{\mathbf{QBS}}_\tau \rightarrow \prob{\mathbf{QBS}}_\tau$ with action on objects
\begin{align*}
    F_q(A) &= \begin{cases}
        \mathbb{R}^D & \text{if } A=\mathbb{R}^D \\
        A \odot A & \text{otherwise},
    \end{cases}
\end{align*}
and action $F_q(f_\theta) : \prob{\mathbf{QBS}}_{\tau}(A, B) \rightarrow \prob{\mathbf{QBS}}_{\tau}(F_q(B), F_q(A))$ on morphisms $f_\theta: \prob{\mathbf{QBS}}_{\tau}(A, B)$ is $F_q(f_\theta) = f^{\dagger}_{\phi}$. Each $f^\dagger_\phi$ is represented by a neural network with parameters $\phi$. The inference functor ``doubles'' objects in order to produce, for each piece of data and each latent variable, two parameters $\mu, \sigma$ for a Gaussian proposal with the appropriate trace-type.

\begin{figure}[t]
    \centering
    \begin{subfigure}{0.45\columnwidth}
        \centering
        \includegraphics[width=\textwidth]{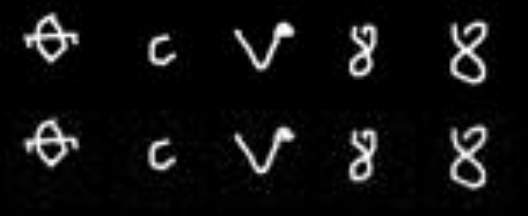}
        \caption{Characters from the Omniglot evaluation set (above) and our model's reconstructions (below)}
        \label{subfig:reconstructions}
    \end{subfigure}
    \hfill
    \begin{subfigure}{0.45\columnwidth}
        \centering
        \includegraphics[width=\textwidth]{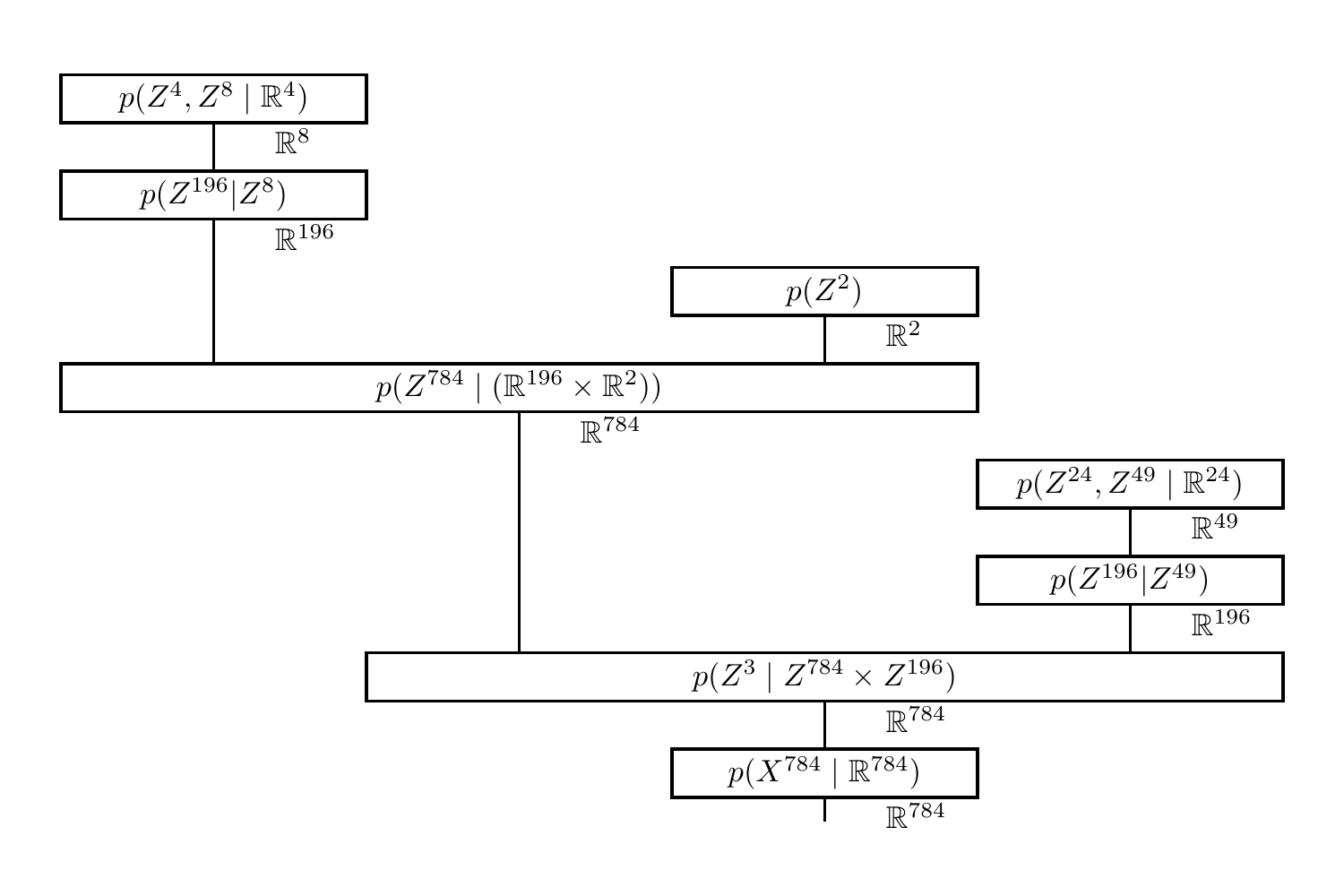}
        \caption{An example string diagram drawn from the approximate posterior of the free category model.}
        \label{subfig:omniglotdiagram}
    \end{subfigure}
    \caption{Reconstructions (left) generated by inference in the diagrammatic generative model (right) on handwritten characters in the Omniglot evaluation set. The string diagram shows a model that generates a glimpse, decodes it into an image canvas via a variational ladder decoder, and then performs a simpler process to generate another glimpse and insert it into the canvas.}
    \label{fig:omniglotresults}
    \vspace{-1em}
\end{figure}

Table~\ref{tab:loglikelihoods} compares the free category prior's performance to other structured generative models (described in Appendix~\ref{app:competing}). We report the estimated log model evidence (estimated as described in Appendix~\ref{app:importanceweighting}). Our free category prior over deep generatives models achieves the best log-evidence per data dimension. Figure~\ref{fig:omniglotresults} shows samples from the trained model's posterior distribution, including reconstruction of evaluation data (Figure~\ref{subfig:reconstructions}) and the morphism sampled for that data (Figure~\ref{subfig:omniglotdiagram}).

\section{Discussion}
\label{sec:discussion}

This paper described a probabilistic generative model over free categories. Section~\ref{sec:categoricalpgm} gave a novel categorical description of distributions over morphisms, by enriching in \textbf{QBS} with joint distributions and push-forwards. It then showed how to sample morphisms from \textbf{QBS}-enriched free monoidal categories via a stochastic short-paths policy. Section~\ref{sec:software} described the software implementation of the free category prior as Discopyro, showing that enriching in a Markov category can also be understood as randomly sampling the computational representations of morphisms in an SMC. It then showed that amortized variational inference in Discopyro achieved competitive log-likelihood in generative modeling of the Omniglot dataset. This section discusses potential future work.

\paragraph{Future improvements to the free category prior}
The underlying quiver $Q$ could be expanded stochastically by randomly choosing and applying commuting diagrams (such as functors, universal constructions, etc.). In the infinite limit, such expansions would give a $Q$ encoding the ``true'' category given by the original graph and the commuting diagrams: a (nonparametric) probability model $\mathbb{P}(Q)$ over the category. This would provide a nonparametric category model as an alternative to nonparametric grammar (i.e.~operad) models~\cite{johnson2006adaptor}. Stochastic memoization~\cite{roy2008stochastic} would enable sampling from that model, with Russian Roulette methods \cite{Xu2019} providing gradient estimates of Equation~\ref{eq:elbo}.

We defined the free category prior in terms of Estrada and Hatano's~\cite{Estrada2008} communicability quasimetric. Boyd et al.~\cite{Boyd2021} recently published a proper metric on quivers and their Markov chains, which might provide a more interpretable foundation for our stochastic short-paths algorithm. Stochastically generalizing shortest-path algorithms may improve the tractability of credit-assignment in policy learning. We plan to explore a subgoal decomposition (e.g.~such as Jurgenson's\cite{Jurgenson2020}) of path-sampling.

\paragraph{Future improvements to the Discopyro implementation}
Discopyro employs amortized variational inference to model the posterior distribution over morphisms. The inference functor mentioned in Section~\ref{subsec:amortizedinf} was defined ad-hoc, meeting only the requirements for constructing faithful inverses \cite{Webb2018} for string diagrams. We used \textbf{QBS} as a semantic category for our learned probabilistic programs, and it is not yet known whether \textbf{QBS} and \textbf{QBS} have all the conditionals necessary to encode Bayesian inversion as a dagger functor. Conjecture~\ref{con:QBSbayes} hypothesizes explicitly that an inference construction similar to ours could approximate Bayesian inversions (in the sense of Cho and Jacobs~\cite{Cho2019}), rather than just a proposal.
\begin{conjecture}[Approximate Bayesian inversions in $\prob{\mathbf{QBS}}$]
\label{con:QBSbayes}
Consider an inference functor $F^{\dagger}$ whose action on objects is identity, with action on morphisms similar to $F_q$ above. Such a functor is a dagger endofunctor $F^{\dagger}: \prob{\mathbf{QBS}} \rightarrow \prob{\mathbf{QBS}}$ sending objects to themselves and morphisms $f: \mathbf{QBS}(A, \prob{B})$ to (approximate) Bayesian inversions $f^\dagger: \mathbf{QBS}(B, \prob{A})$.
\end{conjecture}

\paragraph{Conclusion}
In our experiment the free category prior learned compositional structures from real data. Future applications could include learning program-like representations, a long-term goal for artificial intelligence \cite{Chollet2019,Lake2019b,Nie2020}, or ``cognitive maps'' for spatial navigation \cite{Master2021} in cognitive science \cite{Momennejad2020}.

\nocite{*}
\bibliographystyle{eptcs}
\bibliography{main}

\begin{thebibliography}{10}
\providecommand{\bibitemdeclare}[2]{}
\providecommand{\surnamestart}{}
\providecommand{\surnameend}{}
\providecommand{\urlprefix}{Available at }
\providecommand{\url}[1]{\texttt{#1}}
\providecommand{\href}[2]{\texttt{#2}}
\providecommand{\urlalt}[2]{\href{#1}{#2}}
\providecommand{\doi}[1]{doi:\urlalt{http://dx.doi.org/#1}{#1}}
\providecommand{\bibinfo}[2]{#2}

\bibitemdeclare{article}{baram2018intuitive}
\bibitem{baram2018intuitive}
\bibinfo{author}{Alon~B \surnamestart Baram\surnameend},
  \bibinfo{author}{Timothy~H \surnamestart Muller\surnameend},
  \bibinfo{author}{James~CR \surnamestart Whittington\surnameend} \&
  \bibinfo{author}{Timothy~EJ \surnamestart Behrens\surnameend}
  (\bibinfo{year}{2018}): \emph{\bibinfo{title}{Intuitive planning: global
  navigation through cognitive maps based on grid-like codes}}.
\newblock {\sl \bibinfo{journal}{bioRxiv}}, p. \bibinfo{pages}{421461}.

\bibitemdeclare{article}{bingham2019pyro}
\bibitem{bingham2019pyro}
\bibinfo{author}{Eli \surnamestart Bingham\surnameend},
  \bibinfo{author}{Jonathan~P \surnamestart Chen\surnameend},
  \bibinfo{author}{Martin \surnamestart Jankowiak\surnameend},
  \bibinfo{author}{Fritz \surnamestart Obermeyer\surnameend},
  \bibinfo{author}{Neeraj \surnamestart Pradhan\surnameend},
  \bibinfo{author}{Theofanis \surnamestart Karaletsos\surnameend},
  \bibinfo{author}{Rohit \surnamestart Singh\surnameend}, \bibinfo{author}{Paul
  \surnamestart Szerlip\surnameend}, \bibinfo{author}{Paul \surnamestart
  Horsfall\surnameend} \& \bibinfo{author}{Noah~D \surnamestart
  Goodman\surnameend} (\bibinfo{year}{2019}): \emph{\bibinfo{title}{Pyro: Deep
  universal probabilistic programming}}.
\newblock {\sl \bibinfo{journal}{The Journal of Machine Learning Research}}
  \bibinfo{volume}{20}(\bibinfo{number}{1}), pp. \bibinfo{pages}{973--978}.

\bibitemdeclare{misc}{Blei2016}
\bibitem{Blei2016}
\bibinfo{author}{David~M. \surnamestart Blei\surnameend}, \bibinfo{author}{Alp
  \surnamestart Kucukelbir\surnameend} \& \bibinfo{author}{Jon~D. \surnamestart
  McAuliffe\surnameend} (\bibinfo{year}{2017}):
  \emph{\bibinfo{title}{{Variational Inference: A Review for Statisticians}}},
  \doi{10.1080/01621459.2017.1285773}.
\newblock \urlprefix\url{http://arxiv.org/abs/1601.00670
  http://dx.doi.org/10.1080/01621459.2017.1285773}.

\bibitemdeclare{article}{Boyd2021}
\bibitem{Boyd2021}
\bibinfo{author}{Zachary~M. \surnamestart Boyd\surnameend},
  \bibinfo{author}{Nicolas \surnamestart Fraiman\surnameend},
  \bibinfo{author}{Jeremy \surnamestart Marzuola\surnameend},
  \bibinfo{author}{Peter~J. \surnamestart Mucha\surnameend},
  \bibinfo{author}{Braxton \surnamestart Osting\surnameend} \&
  \bibinfo{author}{Jonathan \surnamestart Weare\surnameend}
  (\bibinfo{year}{2021}): \emph{\bibinfo{title}{{A Metric on Directed Graphs
  and Markov Chains Based on Hitting Probabilities}}}.
\newblock {\sl \bibinfo{journal}{SIAM Journal on Mathematics of Data Science}}
  \bibinfo{volume}{3}(\bibinfo{number}{2}), pp. \bibinfo{pages}{467--493},
  \doi{10.1137/20m1348315}.

\bibitemdeclare{article}{Brualdi1980}
\bibitem{Brualdi1980}
\bibinfo{author}{Richard~A. \surnamestart Brualdi\surnameend},
  \bibinfo{author}{Frank \surnamestart Harary\surnameend} \&
  \bibinfo{author}{Zevi \surnamestart Miller\surnameend}
  (\bibinfo{year}{1980}): \emph{\bibinfo{title}{Bigraphs versus digraphs via
  matrices}}.
\newblock {\sl \bibinfo{journal}{Journal of Graph Theory}}
  \bibinfo{volume}{4}(\bibinfo{number}{1}), pp. \bibinfo{pages}{51--73},
  \doi{https://doi.org/10.1002/jgt.3190040107}.
\newblock
  \urlprefix\url{https://onlinelibrary.wiley.com/doi/abs/10.1002/jgt.3190040107}.

\bibitemdeclare{article}{Cho2019}
\bibitem{Cho2019}
\bibinfo{author}{Kenta \surnamestart Cho\surnameend} \& \bibinfo{author}{Bart
  \surnamestart Jacobs\surnameend} (\bibinfo{year}{2019}):
  \emph{\bibinfo{title}{{Disintegration and Bayesian inversion via string
  diagrams}}}.
\newblock {\sl \bibinfo{journal}{Mathematical Structures in Computer Science}}
  \bibinfo{volume}{29}(\bibinfo{number}{7}), pp. \bibinfo{pages}{938--971},
  \doi{10.1017/S0960129518000488}.

\bibitemdeclare{article}{Chollet2019}
\bibitem{Chollet2019}
\bibinfo{author}{Fran{\c{c}}ois \surnamestart Chollet\surnameend}
  (\bibinfo{year}{2019}): \emph{\bibinfo{title}{{On the Measure of
  Intelligence}}}, pp. \bibinfo{pages}{1--64}.
\newblock \urlprefix\url{http://arxiv.org/abs/1911.01547}.

\bibitemdeclare{book}{Chopin2020}
\bibitem{Chopin2020}
\bibinfo{author}{Nicolas \surnamestart Chopin\surnameend} \&
  \bibinfo{author}{Omiros \surnamestart Papaspiliopoulos\surnameend}
  (\bibinfo{year}{2020}): \emph{\bibinfo{title}{{An Introduction to Sequential
  Monte Carlo}}}.
\newblock \bibinfo{publisher}{Springer}, \doi{10.1007/978-3-030-47845-2}.

\bibitemdeclare{inproceedings}{Clark2008}
\bibitem{Clark2008}
\bibinfo{author}{Stephen \surnamestart Clark\surnameend}, \bibinfo{author}{Bob
  \surnamestart Coecke\surnameend} \& \bibinfo{author}{Mehrnoosh \surnamestart
  Sadrzadeh\surnameend} (\bibinfo{year}{2008}): \emph{\bibinfo{title}{{A
  compositional distributional model of meaning}}}.
\newblock In: {\sl \bibinfo{booktitle}{Proceedings of the Second Quantum
  Interaction Symposium (QI-2008)}}, \bibinfo{volume}{Schuetze 1998}, pp.
  \bibinfo{pages}{133--140}.

\bibitemdeclare{inproceedings}{Cruttwell2021}
\bibitem{Cruttwell2021}
\bibinfo{author}{G.~S.~H. \surnamestart Cruttwell\surnameend},
  \bibinfo{author}{Bruno \surnamestart Gavranovi{\'{c}}\surnameend},
  \bibinfo{author}{Neil \surnamestart Ghani\surnameend}, \bibinfo{author}{Paul
  \surnamestart Wilson\surnameend} \& \bibinfo{author}{Fabio \surnamestart
  Zanasi\surnameend} (\bibinfo{year}{2021}): \emph{\bibinfo{title}{{Categorical
  Foundations of Gradient-Based Learning}}}.
\newblock In: {\sl \bibinfo{booktitle}{Applied Category Theory Conference (ACT
  2021)}}.
\newblock \urlprefix\url{http://arxiv.org/abs/2103.01931}.

\bibitemdeclare{article}{Culbertson2014}
\bibitem{Culbertson2014}
\bibinfo{author}{Jared \surnamestart Culbertson\surnameend} \&
  \bibinfo{author}{Kirk \surnamestart Sturtz\surnameend}
  (\bibinfo{year}{2014}): \emph{\bibinfo{title}{{A categorical foundation for
  bayesian probability}}}.
\newblock {\sl \bibinfo{journal}{Applied Categorical Structures}}
  \bibinfo{volume}{22}(\bibinfo{number}{4}), pp. \bibinfo{pages}{647--662},
  \doi{10.1007/s10485-013-9324-9}.

\bibitemdeclare{article}{BorelKernels}
\bibitem{BorelKernels}
\bibinfo{author}{Fredrik \surnamestart Dahlqvist\surnameend},
  \bibinfo{author}{Alexandra \surnamestart Silva\surnameend},
  \bibinfo{author}{Vincent \surnamestart Danos\surnameend} \&
  \bibinfo{author}{Ilias \surnamestart Garnier\surnameend}
  (\bibinfo{year}{2018}): \emph{\bibinfo{title}{Borel Kernels and their
  Approximation, Categorically}}.
\newblock {\sl \bibinfo{journal}{Electronic Notes in Theoretical Computer
  Science}} \bibinfo{volume}{341}, pp. \bibinfo{pages}{91--119},
  \doi{https://doi.org/10.1016/j.entcs.2018.11.006}.
\newblock
  \urlprefix\url{https://www.sciencedirect.com/science/article/pii/S1571066118300860}.
\newblock \bibinfo{note}{Proceedings of the Thirty-Fourth Conference on the
  Mathematical Foundations of Programming Semantics (MFPS XXXIV)}.

\bibitemdeclare{article}{Elliott2017}
\bibitem{Elliott2017}
\bibinfo{author}{Conal \surnamestart Elliott\surnameend}
  (\bibinfo{year}{2017}): \emph{\bibinfo{title}{{Compiling to categories}}}.
\newblock {\sl \bibinfo{journal}{Proceedings of the ACM on Programming
  Languages}} \bibinfo{volume}{1}(\bibinfo{number}{ICFP}), pp.
  \bibinfo{pages}{1--27}, \doi{10.1145/3110271}.

\bibitemdeclare{article}{Estrada2008}
\bibitem{Estrada2008}
\bibinfo{author}{Ernesto \surnamestart Estrada\surnameend} \&
  \bibinfo{author}{Naomichi \surnamestart Hatano\surnameend}
  (\bibinfo{year}{2008}): \emph{\bibinfo{title}{{Communicability in complex
  networks}}}.
\newblock {\sl \bibinfo{journal}{Physical Review E - Statistical, Nonlinear,
  and Soft Matter Physics}} \bibinfo{volume}{77}(\bibinfo{number}{3}), pp.
  \bibinfo{pages}{1--12}, \doi{10.1103/PhysRevE.77.036111}.

\bibitemdeclare{inproceedings}{Feinman2020}
\bibitem{Feinman2020}
\bibinfo{author}{Reuben \surnamestart Feinman\surnameend} \&
  \bibinfo{author}{Brenden~M. \surnamestart Lake\surnameend}
  (\bibinfo{year}{2021}): \emph{\bibinfo{title}{{Learning Task-General
  Representations with Generative Neuro-Symbolic Modeling}}}.
\newblock In: {\sl \bibinfo{booktitle}{International Conference on Learning
  Representations}}.
\newblock \urlprefix\url{http://arxiv.org/abs/2006.14448}.

\bibitemdeclare{inproceedings}{DeFelice2020}
\bibitem{DeFelice2020}
\bibinfo{author}{Giovanni \surnamestart de~Felice\surnameend},
  \bibinfo{author}{Alexis \surnamestart Toumi\surnameend} \&
  \bibinfo{author}{Bob \surnamestart Coecke\surnameend} (\bibinfo{year}{2020}):
  \emph{\bibinfo{title}{{DisCoPy: Monoidal Categories in Python}}}.
\newblock In: {\sl \bibinfo{booktitle}{Applied Category Theory Conference}},
  pp. \bibinfo{pages}{1--20}.
\newblock \urlprefix\url{http://arxiv.org/abs/2005.02975}.

\bibitemdeclare{article}{Fong2019}
\bibitem{Fong2019}
\bibinfo{author}{Brendan \surnamestart Fong\surnameend} \&
  \bibinfo{author}{Michael \surnamestart Johnson\surnameend}
  (\bibinfo{year}{2019}): \emph{\bibinfo{title}{{Lenses and learners}}}.
\newblock {\sl \bibinfo{journal}{CEUR Workshop Proceedings}}
  \bibinfo{volume}{2355}(\bibinfo{number}{Bx}), pp. \bibinfo{pages}{16--29}.

\bibitemdeclare{article}{Fong2019a}
\bibitem{Fong2019a}
\bibinfo{author}{Brendan \surnamestart Fong\surnameend}, \bibinfo{author}{David
  \surnamestart Spivak\surnameend} \& \bibinfo{author}{Remy \surnamestart
  Tuyeras\surnameend} (\bibinfo{year}{2019}): \emph{\bibinfo{title}{{Backprop
  as Functor: A compositional perspective on supervised learning}}}.
\newblock {\sl \bibinfo{journal}{Proceedings - Symposium on Logic in Computer
  Science}} \bibinfo{volume}{2019-June}, pp. \bibinfo{pages}{1--13},
  \doi{10.1109/LICS.2019.8785665}.

\bibitemdeclare{book}{Fong2019b}
\bibitem{Fong2019b}
\bibinfo{author}{Brendan \surnamestart Fong\surnameend} \&
  \bibinfo{author}{David~I \surnamestart Spivak\surnameend}
  (\bibinfo{year}{2019}): \emph{\bibinfo{title}{{Seven Sketches in
  Compositionality: An Invitation to Applied Category Theory}}}.
\newblock \bibinfo{publisher}{Cambridge University Press}.
\newblock \urlprefix\url{http://arxiv.org/abs/1803.05316}.

\bibitemdeclare{article}{Frankland2020}
\bibitem{Frankland2020}
\bibinfo{author}{Steven~M. \surnamestart Frankland\surnameend} \&
  \bibinfo{author}{Joshua~D. \surnamestart Greene\surnameend}
  (\bibinfo{year}{2020}): \emph{\bibinfo{title}{{Concepts and Compositionality:
  In Search of the Brain's Language of Thought}}}.
\newblock {\sl \bibinfo{journal}{Annual Review of Psychology}}
  \bibinfo{volume}{71}(\bibinfo{number}{1}), pp. \bibinfo{pages}{273--303},
  \doi{10.1146/annurev-psych-122216-011829}.

\bibitemdeclare{article}{Fritz2020}
\bibitem{Fritz2020}
\bibinfo{author}{Tobias \surnamestart Fritz\surnameend} (\bibinfo{year}{2020}):
  \emph{\bibinfo{title}{{A synthetic approach to Markov kernels, conditional
  independence and theorems on sufficient statistics}}}.
\newblock {\sl \bibinfo{journal}{Advances in Mathematics}}
  \bibinfo{volume}{370}, p. \bibinfo{pages}{107239},
  \doi{10.1016/j.aim.2020.107239}.
\newblock \urlprefix\url{https://doi.org/10.1016/j.aim.2020.107239}.

\bibitemdeclare{book}{gardenfors2004conceptual}
\bibitem{gardenfors2004conceptual}
\bibinfo{author}{Peter \surnamestart Gardenfors\surnameend}
  (\bibinfo{year}{2004}): \emph{\bibinfo{title}{Conceptual spaces: The geometry
  of thought}}.
\newblock \bibinfo{publisher}{MIT press}.

\bibitemdeclare{inproceedings}{Gavranovic2019}
\bibitem{Gavranovic2019}
\bibinfo{author}{Bruno \surnamestart Gavranovic\surnameend}
  (\bibinfo{year}{2019}): \emph{\bibinfo{title}{{Learning functors using
  gradient descent}}}.
\newblock In: {\sl \bibinfo{booktitle}{Applied Category Theory Conference (ACT
  2019)}}, \bibinfo{volume}{323}, \bibinfo{publisher}{Electronic Proceedings in
  Theoretical Computer Science, EPTCS}, pp. \bibinfo{pages}{230--245},
  \doi{10.4204/EPTCS.323.15}.

\bibitemdeclare{incollection}{giry1982categorical}
\bibitem{giry1982categorical}
\bibinfo{author}{Michele \surnamestart Giry\surnameend} (\bibinfo{year}{1982}):
  \emph{\bibinfo{title}{A categorical approach to probability theory}}.
\newblock In: {\sl \bibinfo{booktitle}{Categorical aspects of topology and
  analysis}}, \bibinfo{publisher}{Springer}, pp. \bibinfo{pages}{68--85}.

\bibitemdeclare{article}{golubtsov1999axiomatic}
\bibitem{golubtsov1999axiomatic}
\bibinfo{author}{Petr~Viktorovich \surnamestart Golubtsov\surnameend}
  (\bibinfo{year}{1999}): \emph{\bibinfo{title}{Axiomatic description of
  categories of information transformers}}.
\newblock {\sl \bibinfo{journal}{Problemy Peredachi Informatsii}}
  \bibinfo{volume}{35}(\bibinfo{number}{3}), pp. \bibinfo{pages}{80--98}.

\bibitemdeclare{inproceedings}{Grant2019}
\bibitem{Grant2019}
\bibinfo{author}{Erin \surnamestart Grant\surnameend},
  \bibinfo{author}{Joshua~C. \surnamestart Peterson\surnameend} \&
  \bibinfo{author}{Tom \surnamestart Griffiths\surnameend}
  (\bibinfo{year}{2019}): \emph{\bibinfo{title}{{Learning deep taxonomic priors
  for concept learning from few positive examples}}}.
\newblock In: {\sl \bibinfo{booktitle}{The Annual Meeting of the Cognitive
  Science Society}}, pp. \bibinfo{pages}{1865--1870}.
\newblock
  \urlprefix\url{https://www.semanticscholar.org/paper/Learning-deep-taxonomic-priors-for-concept-learning-Grant-Peterson/8eef236bca7ed58f8fa96786925c0e1da4a124eb}.

\bibitemdeclare{inproceedings}{Grefenstette2011}
\bibitem{Grefenstette2011}
\bibinfo{author}{Edward \surnamestart Grefenstette\surnameend} \&
  \bibinfo{author}{Mehrnoosh \surnamestart Sadrzadeh\surnameend}
  (\bibinfo{year}{2011}): \emph{\bibinfo{title}{{Experimental support for a
  categorical compositional distributional model of meaning}}}.
\newblock In: {\sl \bibinfo{booktitle}{EMNLP 2011 - Conference on Empirical
  Methods in Natural Language Processing, Proceedings of the Conference}}, pp.
  \bibinfo{pages}{1394--1404}.

\bibitemdeclare{article}{Halter2020}
\bibitem{Halter2020}
\bibinfo{author}{Micah \surnamestart Halter\surnameend}, \bibinfo{author}{Evan
  \surnamestart Patterson\surnameend}, \bibinfo{author}{Andrew \surnamestart
  Baas\surnameend} \& \bibinfo{author}{James \surnamestart
  Fairbanks\surnameend} (\bibinfo{year}{2020}):
  \emph{\bibinfo{title}{{Compositional Scientific Computing with Catlab and
  SemanticModels}}}, pp. \bibinfo{pages}{1--3}.
\newblock \urlprefix\url{http://arxiv.org/abs/2005.04831}.

\bibitemdeclare{article}{Hasson2020}
\bibitem{Hasson2020}
\bibinfo{author}{Uri \surnamestart Hasson\surnameend},
  \bibinfo{author}{Samuel~A. \surnamestart Nastase\surnameend} \&
  \bibinfo{author}{Ariel \surnamestart Goldstein\surnameend}
  (\bibinfo{year}{2020}): \emph{\bibinfo{title}{{Direct Fit to Nature: An
  Evolutionary Perspective on Biological and Artificial Neural Networks}}}.
\newblock {\sl \bibinfo{journal}{Neuron}}
  \bibinfo{volume}{105}(\bibinfo{number}{3}), pp. \bibinfo{pages}{416--434},
  \doi{10.1016/j.neuron.2019.12.002}.
\newblock \urlprefix\url{https://doi.org/10.1016/j.neuron.2019.12.002}.

\bibitemdeclare{inproceedings}{He2019}
\bibitem{He2019}
\bibinfo{author}{Jiawei \surnamestart He\surnameend},
  \bibinfo{author}{Yu~\surnamestart Gong\surnameend}, \bibinfo{author}{Greg
  \surnamestart Mori\surnameend}, \bibinfo{author}{Joseph \surnamestart
  Marino\surnameend} \& \bibinfo{author}{Andreas~M. \surnamestart
  Lehrmann\surnameend} (\bibinfo{year}{2019}):
  \emph{\bibinfo{title}{{Variational autoencoders with jointly optimized latent
  dependency structure}}}.
\newblock In: {\sl \bibinfo{booktitle}{7th International Conference on Learning
  Representations, ICLR 2019}}, pp. \bibinfo{pages}{1--16}.

\bibitemdeclare{inproceedings}{Hermida1998}
\bibitem{Hermida1998}
\bibinfo{author}{C.~\surnamestart Hermida\surnameend},
  \bibinfo{author}{M.~\surnamestart Makkai\surnameend} \&
  \bibinfo{author}{J.~\surnamestart Power\surnameend} (\bibinfo{year}{1998}):
  \emph{\bibinfo{title}{Higher dimensional multigraphs}}.
\newblock In: {\sl \bibinfo{booktitle}{Proceedings. Thirteenth Annual IEEE
  Symposium on Logic in Computer Science (Cat. No.98CB36226)}}, pp.
  \bibinfo{pages}{199--206}, \doi{10.1109/LICS.1998.705656}.

\bibitemdeclare{inproceedings}{Heunen2017}
\bibitem{Heunen2017}
\bibinfo{author}{Chris \surnamestart Heunen\surnameend}, \bibinfo{author}{Ohad
  \surnamestart Kammar\surnameend}, \bibinfo{author}{Sam \surnamestart
  Staton\surnameend} \& \bibinfo{author}{Hongseok \surnamestart
  Yang\surnameend} (\bibinfo{year}{2017}): \emph{\bibinfo{title}{{A convenient
  category for higher-order probability theory}}}.
\newblock In: {\sl \bibinfo{booktitle}{Proceedings - Symposium on Logic in
  Computer Science}}, \doi{10.1109/LICS.2017.8005137}.

\bibitemdeclare{article}{Hewitt2018}
\bibitem{Hewitt2018}
\bibinfo{author}{Luke~B. \surnamestart Hewitt\surnameend},
  \bibinfo{author}{Maxwell~I. \surnamestart Nye\surnameend},
  \bibinfo{author}{Andreea \surnamestart Gane\surnameend},
  \bibinfo{author}{Tommi \surnamestart Jaakkola\surnameend} \&
  \bibinfo{author}{Joshua~B. \surnamestart Tenenbaum\surnameend}
  (\bibinfo{year}{2018}): \emph{\bibinfo{title}{{The Variational Homoencoder:
  Learning to learn high capacity generative models from few examples}}}.
\newblock {\sl \bibinfo{journal}{34th Conference on Uncertainty in Artificial
  Intelligence 2018, UAI 2018}} \bibinfo{volume}{2}, pp.
  \bibinfo{pages}{988--997}.

\bibitemdeclare{inproceedings}{Ho2018}
\bibitem{Ho2018}
\bibinfo{author}{Mark~K \surnamestart Ho\surnameend} \& \bibinfo{author}{Tom
  \surnamestart Griffiths\surnameend} (\bibinfo{year}{2018}):
  \emph{\bibinfo{title}{{Human Priors in Hierarchical Program Induction}}}.
\newblock In: {\sl \bibinfo{booktitle}{Cognitive Computational Neuroscience}},
  \bibinfo{volume}{1}.
\newblock \urlprefix\url{http://lightbot.com}.

\bibitemdeclare{article}{johnson2006adaptor}
\bibitem{johnson2006adaptor}
\bibinfo{author}{Mark \surnamestart Johnson\surnameend},
  \bibinfo{author}{Thomas \surnamestart Griffiths\surnameend} \&
  \bibinfo{author}{Sharon \surnamestart Goldwater\surnameend}
  (\bibinfo{year}{2006}): \emph{\bibinfo{title}{Adaptor grammars: A framework
  for specifying compositional nonparametric Bayesian models}}.
\newblock {\sl \bibinfo{journal}{Advances in neural information processing
  systems}} \bibinfo{volume}{19}.

\bibitemdeclare{inproceedings}{Jurgenson2020}
\bibitem{Jurgenson2020}
\bibinfo{author}{Tom \surnamestart Jurgenson\surnameend},
  \bibinfo{author}{Or~\surnamestart Avner\surnameend}, \bibinfo{author}{Edward
  \surnamestart Groshev\surnameend} \& \bibinfo{author}{Aviv \surnamestart
  Tamar\surnameend} (\bibinfo{year}{2020}): \emph{\bibinfo{title}{{Sub-goal
  trees – A framework for goal-based reinforcement learning}}}.
\newblock In: {\sl \bibinfo{booktitle}{37th International Conference on Machine
  Learning, ICML 2020}}, pp. \bibinfo{pages}{5020--5030}.

\bibitemdeclare{article}{Lake2019}
\bibitem{Lake2019}
\bibinfo{author}{Brenden~M. \surnamestart Lake\surnameend} \&
  \bibinfo{author}{Steven~T. \surnamestart Piantadosi\surnameend}
  (\bibinfo{year}{2019}): \emph{\bibinfo{title}{{People infer recursive visual
  concepts from just a few examples}}}.
\newblock {\sl \bibinfo{journal}{Computational Brain \& Behavior}}.
\newblock \urlprefix\url{http://arxiv.org/abs/1904.08034}.

\bibitemdeclare{article}{Lake2015}
\bibitem{Lake2015}
\bibinfo{author}{Brenden~M. \surnamestart Lake\surnameend},
  \bibinfo{author}{Ruslan \surnamestart Salakhutdinov\surnameend} \&
  \bibinfo{author}{Joshua~B. \surnamestart Tenenbaum\surnameend}
  (\bibinfo{year}{2015}): \emph{\bibinfo{title}{{Human-level concept learning
  through probabilistic program induction}}}.
\newblock {\sl \bibinfo{journal}{Science}}
  \bibinfo{volume}{350}(\bibinfo{number}{6266}), pp.
  \bibinfo{pages}{1332--1338}, \doi{10.1126/science.aab3050}.
\newblock \urlprefix\url{http://science.sciencemag.org/content/350/6266/1332
  https://www.sciencemag.org/content/350/6266/1332.full.pdf}.

\bibitemdeclare{article}{Lake2019b}
\bibitem{Lake2019b}
\bibinfo{author}{Brenden~M. \surnamestart Lake\surnameend},
  \bibinfo{author}{Ruslan \surnamestart Salakhutdinov\surnameend} \&
  \bibinfo{author}{Joshua~B. \surnamestart Tenenbaum\surnameend}
  (\bibinfo{year}{2019}): \emph{\bibinfo{title}{{The Omniglot challenge: a
  3-year progress report}}}.
\newblock {\sl \bibinfo{journal}{Current Opinion in Behavioral Sciences}}
  \bibinfo{volume}{29}, pp. \bibinfo{pages}{97--104},
  \doi{10.1016/j.cobeha.2019.04.007}.
\newblock \urlprefix\url{https://doi.org/10.1016/j.cobeha.2019.04.007}.

\bibitemdeclare{book}{Latouche1999}
\bibitem{Latouche1999}
\bibinfo{author}{Guy \surnamestart Latouche\surnameend} \&
  \bibinfo{author}{Vaidyanathan \surnamestart Ramaswami\surnameend}
  (\bibinfo{year}{1999}): \emph{\bibinfo{title}{{Introduction to Matrix
  Analytic Methods in Stochastic Modeling}}}.
\newblock \bibinfo{publisher}{Society for Industrial and Applied Mathematics},
  \bibinfo{address}{Philadelphia, PA}.

\bibitemdeclare{inproceedings}{Lew2020}
\bibitem{Lew2020}
\bibinfo{author}{Alexander~K. \surnamestart Lew\surnameend},
  \bibinfo{author}{Marco~F. \surnamestart Cusumano-Towner\surnameend},
  \bibinfo{author}{Benjamin \surnamestart Sherman\surnameend},
  \bibinfo{author}{Michael \surnamestart Carbin\surnameend} \&
  \bibinfo{author}{Vikash~K. \surnamestart Mansinghka\surnameend}
  (\bibinfo{year}{2020}): \emph{\bibinfo{title}{{Trace Types and Denotational
  Semantics for Sound Programmable Inference in Probabilistic Languages}}}.
\newblock In: {\sl \bibinfo{booktitle}{ACM Principles of Programming
  Languages}}, \bibinfo{volume}{4}, pp. \bibinfo{pages}{1--31},
  \doi{10.1145/3371087}.

\bibitemdeclare{book}{marcus2018algebraic}
\bibitem{marcus2018algebraic}
\bibinfo{author}{Gary~F \surnamestart Marcus\surnameend}
  (\bibinfo{year}{2018}): \emph{\bibinfo{title}{{The algebraic mind:
  Integrating connectionism and cognitive science}}}.
\newblock \bibinfo{publisher}{MIT press}.

\bibitemdeclare{inproceedings}{Master2021}
\bibitem{Master2021}
\bibinfo{author}{Jade \surnamestart Master\surnameend} (\bibinfo{year}{2021}):
  \emph{\bibinfo{title}{{The Open Algebraic Path Problem}}}.
\newblock In \bibinfo{editor}{Fabio \surnamestart Gadducci\surnameend} \&
  \bibinfo{editor}{Alexandra \surnamestart Silva\surnameend}, editors: {\sl
  \bibinfo{booktitle}{9th Conference on Algebra and Coalgebra in Computer
  Science (CALCO 2021)}}, {\sl \bibinfo{series}{Leibniz International
  Proceedings in Informatics (LIPIcs)}} \bibinfo{volume}{211},
  \bibinfo{publisher}{Schloss Dagstuhl -- Leibniz-Zentrum f{\"u}r Informatik},
  \bibinfo{address}{Dagstuhl, Germany}, pp. \bibinfo{pages}{20:1--20:20},
  \doi{10.4230/LIPIcs.CALCO.2021.20}.
\newblock \urlprefix\url{https://drops.dagstuhl.de/opus/volltexte/2021/15375}.

\bibitemdeclare{article}{van2018introduction}
\bibitem{van2018introduction}
\bibinfo{author}{Jan-Willem \surnamestart van~de Meent\surnameend},
  \bibinfo{author}{Brooks \surnamestart Paige\surnameend},
  \bibinfo{author}{Hongseok \surnamestart Yang\surnameend} \&
  \bibinfo{author}{Frank \surnamestart Wood\surnameend} (\bibinfo{year}{2018}):
  \emph{\bibinfo{title}{An introduction to probabilistic programming}}.
\newblock {\sl \bibinfo{journal}{arXiv preprint arXiv:1809.10756}}.

\bibitemdeclare{inproceedings}{Mnih2014}
\bibitem{Mnih2014}
\bibinfo{author}{Andriy \surnamestart Mnih\surnameend} \&
  \bibinfo{author}{Karol \surnamestart Gregor\surnameend}
  (\bibinfo{year}{2014}): \emph{\bibinfo{title}{{Neural variational inference
  and learning in belief networks}}}.
\newblock In: {\sl \bibinfo{booktitle}{31st International Conference on Machine
  Learning, ICML 2014}}, \bibinfo{volume}{5}, pp. \bibinfo{pages}{3800--3809}.

\bibitemdeclare{article}{Momennejad2020}
\bibitem{Momennejad2020}
\bibinfo{author}{Ida \surnamestart Momennejad\surnameend}
  (\bibinfo{year}{2020}): \emph{\bibinfo{title}{{Learning Structures:
  Predictive Representations, Replay, and Generalization}}}.
\newblock {\sl \bibinfo{journal}{Current Opinion in Behavioral Sciences}}
  \bibinfo{volume}{32}, pp. \bibinfo{pages}{155--166},
  \doi{10.1016/j.cobeha.2020.02.017}.
\newblock \urlprefix\url{https://doi.org/10.1016/j.cobeha.2020.02.017}.

\bibitemdeclare{book}{Murphy2012}
\bibitem{Murphy2012}
\bibinfo{author}{P~K \surnamestart Murphy\surnameend} (\bibinfo{year}{2012}):
  \emph{\bibinfo{title}{{Machine Learning: A Probabilistic Perspective}}}.
\newblock \doi{10.1007/SpringerReference_35834}.

\bibitemdeclare{inproceedings}{Nie2020}
\bibitem{Nie2020}
\bibinfo{author}{Weili \surnamestart Nie\surnameend}, \bibinfo{author}{Zhiding
  \surnamestart Yu\surnameend}, \bibinfo{author}{Lei \surnamestart
  Mao\surnameend}, \bibinfo{author}{Ankit~B. \surnamestart Patel\surnameend},
  \bibinfo{author}{Yuke \surnamestart Zhu\surnameend} \&
  \bibinfo{author}{Animashree \surnamestart Anandkumar\surnameend}
  (\bibinfo{year}{2020}): \emph{\bibinfo{title}{{Bongard-LOGO: A New Benchmark
  for Human-Level Concept Learning and Reasoning}}}.
\newblock In: {\sl \bibinfo{booktitle}{Advances in Neural Information
  Processing Systems}}, \bibinfo{volume}{NeurIPS}.
\newblock \urlprefix\url{http://arxiv.org/abs/2010.00763}.

\bibitemdeclare{article}{Nye2020}
\bibitem{Nye2020}
\bibinfo{author}{Maxwell~I. \surnamestart Nye\surnameend},
  \bibinfo{author}{Armando \surnamestart Solar-Lezama\surnameend},
  \bibinfo{author}{Joshua~B. \surnamestart Tenenbaum\surnameend} \&
  \bibinfo{author}{Brenden~M. \surnamestart Lake\surnameend}
  (\bibinfo{year}{2020}): \emph{\bibinfo{title}{{Learning compositional rules
  via neural program synthesis}}}.
\newblock {\sl \bibinfo{journal}{Advances in Neural Information Processing
  Systems}} \bibinfo{volume}{2020-December}(\bibinfo{number}{NeurIPS}), pp.
  \bibinfo{pages}{1--11}.

\bibitemdeclare{article}{Overlan2017}
\bibitem{Overlan2017}
\bibinfo{author}{Matthew~C. \surnamestart Overlan\surnameend},
  \bibinfo{author}{Robert~A. \surnamestart Jacobs\surnameend} \&
  \bibinfo{author}{Steven~T. \surnamestart Piantadosi\surnameend}
  (\bibinfo{year}{2017}): \emph{\bibinfo{title}{{Learning abstract visual
  concepts via probabilistic program induction in a Language of Thought}}}.
\newblock {\sl \bibinfo{journal}{Cognition}} \bibinfo{volume}{168}, pp.
  \bibinfo{pages}{320--334}, \doi{10.1016/j.cognition.2017.07.005}.
\newblock \urlprefix\url{http://dx.doi.org/10.1016/j.cognition.2017.07.005}.

\bibitemdeclare{inproceedings}{Parisotto2017}
\bibitem{Parisotto2017}
\bibinfo{author}{Emilio \surnamestart Parisotto\surnameend},
  \bibinfo{author}{Abdel-rahman \surnamestart Mohamed\surnameend},
  \bibinfo{author}{Rishabh \surnamestart Singh\surnameend},
  \bibinfo{author}{Lihong \surnamestart Li\surnameend},
  \bibinfo{author}{Dengyong \surnamestart Zhou\surnameend} \&
  \bibinfo{author}{Pushmeet \surnamestart Kohli\surnameend}
  (\bibinfo{year}{2017}): \emph{\bibinfo{title}{{Neuro-Symbolic Program
  Synthesis}}}.
\newblock In: {\sl \bibinfo{booktitle}{International Conference on Learning
  Representations}}, pp. \bibinfo{pages}{1--14}.
\newblock \urlprefix\url{http://arxiv.org/abs/1611.01855}.

\bibitemdeclare{article}{Patterson2021b}
\bibitem{Patterson2021b}
\bibinfo{author}{Evan \surnamestart Patterson\surnameend},
  \bibinfo{author}{Owen \surnamestart Lynch\surnameend} \&
  \bibinfo{author}{James \surnamestart Fairbanks\surnameend}
  (\bibinfo{year}{2021}): \emph{\bibinfo{title}{{Categorical Data Structures
  for Technical Computing}}}, pp. \bibinfo{pages}{1--28}.
\newblock \urlprefix\url{https://arxiv.org/abs/2106.04703}.

\bibitemdeclare{article}{Patterson2021}
\bibitem{Patterson2021}
\bibinfo{author}{Evan \surnamestart Patterson\surnameend},
  \bibinfo{author}{David~I. \surnamestart Spivak\surnameend} \&
  \bibinfo{author}{Dmitry \surnamestart Vagner\surnameend}
  (\bibinfo{year}{2021}): \emph{\bibinfo{title}{{Wiring diagrams as normal
  forms for computing in symmetric monoidal categories}}}.
\newblock {\sl \bibinfo{journal}{Electronic Proceedings in Theoretical Computer
  Science, EPTCS}} \bibinfo{volume}{333}, pp. \bibinfo{pages}{49--64},
  \doi{10.4204/EPTCS.333.4}.

\bibitemdeclare{article}{Phillips2010}
\bibitem{Phillips2010}
\bibinfo{author}{Steven \surnamestart Phillips\surnameend} \&
  \bibinfo{author}{William~H. \surnamestart Wilson\surnameend}
  (\bibinfo{year}{2010}): \emph{\bibinfo{title}{Categorial Compositionality: A
  Category Theory Explanation for the Systematicity of Human Cognition}}.
\newblock {\sl \bibinfo{journal}{PLOS Computational Biology}}
  \bibinfo{volume}{6}(\bibinfo{number}{7}), pp. \bibinfo{pages}{1--14},
  \doi{10.1371/journal.pcbi.1000858}.
\newblock \urlprefix\url{https://doi.org/10.1371/journal.pcbi.1000858}.

\bibitemdeclare{article}{Piantadosi2016a}
\bibitem{Piantadosi2016a}
\bibinfo{author}{Steven~T. \surnamestart Piantadosi\surnameend},
  \bibinfo{author}{Joshua~B. \surnamestart Tenenbaum\surnameend} \&
  \bibinfo{author}{Noah~D. \surnamestart Goodman\surnameend}
  (\bibinfo{year}{2016}): \emph{\bibinfo{title}{{The logical primitives of
  thought: Empirical foundations for compositional cognitive models}}}.
\newblock {\sl \bibinfo{journal}{Psychological Review}}
  \bibinfo{volume}{123}(\bibinfo{number}{4}), pp. \bibinfo{pages}{392--424},
  \doi{10.1037/a0039980}.

\bibitemdeclare{inproceedings}{Rezende2016}
\bibitem{Rezende2016}
\bibinfo{author}{Danilo~Jimenez \surnamestart Rezende\surnameend},
  \bibinfo{author}{Shakir \surnamestart Mohamed\surnameend},
  \bibinfo{author}{Ivo \surnamestart Danihelka\surnameend},
  \bibinfo{author}{Karol \surnamestart Gregor\surnameend} \&
  \bibinfo{author}{Daan \surnamestart Wierstra\surnameend}
  (\bibinfo{year}{2016}): \emph{\bibinfo{title}{{One-Shot Generalization in
  Deep Generative Models}}}.
\newblock In: {\sl \bibinfo{booktitle}{International Conference on Machine
  Learning}}, \bibinfo{volume}{48}.

\bibitemdeclare{article}{Romano2018}
\bibitem{Romano2018}
\bibinfo{author}{Sergio \surnamestart Romano\surnameend},
  \bibinfo{author}{Alejo \surnamestart Salles\surnameend},
  \bibinfo{author}{Marie \surnamestart Amalric\surnameend},
  \bibinfo{author}{Stanislas \surnamestart Dehaene\surnameend},
  \bibinfo{author}{Mariano \surnamestart Sigman\surnameend} \&
  \bibinfo{author}{Santiago \surnamestart Figueira\surnameend}
  (\bibinfo{year}{2018}): \emph{\bibinfo{title}{{Bayesian validation of grammar
  productions for the language of thought}}}.
\newblock {\sl \bibinfo{journal}{PLoS ONE}}
  \bibinfo{volume}{13}(\bibinfo{number}{7}), pp. \bibinfo{pages}{1--20},
  \doi{10.1371/journal.pone.0200420}.

\bibitemdeclare{inproceedings}{roy2008stochastic}
\bibitem{roy2008stochastic}
\bibinfo{author}{Daniel~M \surnamestart Roy\surnameend},
  \bibinfo{author}{VK~\surnamestart Mansinghka\surnameend},
  \bibinfo{author}{ND~\surnamestart Goodman\surnameend} \&
  \bibinfo{author}{JB~\surnamestart Tenenbaum\surnameend}
  (\bibinfo{year}{2008}): \emph{\bibinfo{title}{A stochastic programming
  perspective on nonparametric Bayes}}.
\newblock In: {\sl \bibinfo{booktitle}{Nonparametric Bayesian Workshop, Int.
  Conf. on Machine Learning}}, \bibinfo{volume}{22}, p.~\bibinfo{pages}{26}.

\bibitemdeclare{article}{Rupel2013}
\bibitem{Rupel2013}
\bibinfo{author}{Dylan \surnamestart Rupel\surnameend} \&
  \bibinfo{author}{David~I. \surnamestart Spivak\surnameend}
  (\bibinfo{year}{2013}): \emph{\bibinfo{title}{{The operad of temporal wiring
  diagrams: formalizing a graphical language for discrete-time processes}}},
  pp. \bibinfo{pages}{1--37}.
\newblock \urlprefix\url{http://arxiv.org/abs/1307.6894}.

\bibitemdeclare{inproceedings}{Schulman2015}
\bibitem{Schulman2015}
\bibinfo{author}{John \surnamestart Schulman\surnameend},
  \bibinfo{author}{Nicolas \surnamestart Heess\surnameend},
  \bibinfo{author}{Theophane \surnamestart Weber\surnameend} \&
  \bibinfo{author}{Pieter \surnamestart Abbeel\surnameend}
  (\bibinfo{year}{2015}): \emph{\bibinfo{title}{{Gradient estimation using
  stochastic computation graphs}}}.
\newblock In: {\sl \bibinfo{booktitle}{Advances in Neural Information
  Processing Systems}}, \bibinfo{volume}{2015-Janua}, pp.
  \bibinfo{pages}{3528--3536}.

\bibitemdeclare{inproceedings}{Scibior2018}
\bibitem{Scibior2018}
\bibinfo{author}{Adam \surnamestart {\'{S}}cibior\surnameend},
  \bibinfo{author}{Ohad \surnamestart Kammar\surnameend},
  \bibinfo{author}{Matthijs \surnamestart V{\'{a}}k{\'{a}}r\surnameend},
  \bibinfo{author}{Sam \surnamestart Staton\surnameend},
  \bibinfo{author}{Hongseok \surnamestart Yang\surnameend},
  \bibinfo{author}{Yufei \surnamestart Cai\surnameend}, \bibinfo{author}{Klaus
  \surnamestart Ostermann\surnameend}, \bibinfo{author}{Sean~K. \surnamestart
  Moss\surnameend}, \bibinfo{author}{Chris \surnamestart Heunen\surnameend} \&
  \bibinfo{author}{Zoubin \surnamestart Ghahramani\surnameend}
  (\bibinfo{year}{2018}): \emph{\bibinfo{title}{{Denotational validation of
  higher-order Bayesian inference}}}.
\newblock In: {\sl \bibinfo{booktitle}{Principles of Programming Languages}},
  \bibinfo{volume}{2}, \doi{10.1145/3158148}.

\bibitemdeclare{inproceedings}{Shiebler2021}
\bibitem{Shiebler2021}
\bibinfo{author}{Dan \surnamestart Shiebler\surnameend}, \bibinfo{author}{Bruno
  \surnamestart Gavranovi{\'{c}}\surnameend} \& \bibinfo{author}{Paul
  \surnamestart Wilson\surnameend} (\bibinfo{year}{2021}):
  \emph{\bibinfo{title}{{Category Theory in Machine Learning}}}.
\newblock In: {\sl \bibinfo{booktitle}{Applied Category Theory Conference}}.
\newblock \urlprefix\url{http://arxiv.org/abs/2106.07032}.

\bibitemdeclare{article}{SOLOMONOFF19641}
\bibitem{SOLOMONOFF19641}
\bibinfo{author}{R~J \surnamestart Solomonoff\surnameend}
  (\bibinfo{year}{1964}): \emph{\bibinfo{title}{{A formal theory of inductive
  inference. Part I}}}.
\newblock {\sl \bibinfo{journal}{Information and Control}}
  \bibinfo{volume}{7}(\bibinfo{number}{1}), pp. \bibinfo{pages}{1--22},
  \doi{https://doi.org/10.1016/S0019-9958(64)90223-2}.
\newblock
  \urlprefix\url{http://www.sciencedirect.com/science/article/pii/S0019995864902232}.

\bibitemdeclare{book}{sorensen2006lectures}
\bibitem{sorensen2006lectures}
\bibinfo{author}{Morten~Heine \surnamestart S{\o}rensen\surnameend} \&
  \bibinfo{author}{Pawel \surnamestart Urzyczyn\surnameend}
  (\bibinfo{year}{2006}): \emph{\bibinfo{title}{Lectures on the Curry-Howard
  isomorphism}}.
\newblock \bibinfo{publisher}{Elsevier}.

\bibitemdeclare{article}{Spivak2013}
\bibitem{Spivak2013}
\bibinfo{author}{David~I \surnamestart Spivak\surnameend}
  (\bibinfo{year}{2013}): \emph{\bibinfo{title}{The operad of wiring diagrams:
  formalizing a graphical language for databases, recursion, and plug-and-play
  circuits}}.
\newblock {\sl \bibinfo{journal}{arXiv preprint arXiv:1305.0297}}.

\bibitemdeclare{inproceedings}{Stuhlmuller2013}
\bibitem{Stuhlmuller2013}
\bibinfo{author}{Andreas \surnamestart Stuhlm\"{u}ller\surnameend},
  \bibinfo{author}{Jacob \surnamestart Taylor\surnameend} \&
  \bibinfo{author}{Noah \surnamestart Goodman\surnameend}
  (\bibinfo{year}{2013}): \emph{\bibinfo{title}{Learning Stochastic Inverses}}.
\newblock In \bibinfo{editor}{C.~J.~C. \surnamestart Burges\surnameend},
  \bibinfo{editor}{L.~\surnamestart Bottou\surnameend},
  \bibinfo{editor}{M.~\surnamestart Welling\surnameend},
  \bibinfo{editor}{Z.~\surnamestart Ghahramani\surnameend} \&
  \bibinfo{editor}{K.~Q. \surnamestart Weinberger\surnameend}, editors: {\sl
  \bibinfo{booktitle}{Advances in Neural Information Processing Systems}},
  \bibinfo{volume}{26}, \bibinfo{publisher}{Curran Associates, Inc.}
\newblock
  \urlprefix\url{https://proceedings.neurips.cc/paper/2013/file/7f53f8c6c730af6aeb52e66eb74d8507-Paper.pdf}.

\bibitemdeclare{inproceedings}{Valkov2018}
\bibitem{Valkov2018}
\bibinfo{author}{Lazar \surnamestart Valkov\surnameend}, \bibinfo{author}{Dipak
  \surnamestart Chaudhari\surnameend}, \bibinfo{author}{Charles \surnamestart
  Sutton\surnameend}, \bibinfo{author}{Akash \surnamestart
  Srivastava\surnameend} \& \bibinfo{author}{Swarat \surnamestart
  Chaudhuri\surnameend} (\bibinfo{year}{2018}): \emph{\bibinfo{title}{{Houdini:
  Lifelong learning as program synthesis}}}.
\newblock In: {\sl \bibinfo{booktitle}{Advances in Neural Information
  Processing Systems}}, \bibinfo{volume}{2018-Decem}, pp.
  \bibinfo{pages}{8687--8698}.

\bibitemdeclare{article}{Vitanyi2000}
\bibitem{Vitanyi2000}
\bibinfo{author}{Paul~M.B. \surnamestart Vit{\'{a}}nyi\surnameend} \&
  \bibinfo{author}{Ming \surnamestart Li\surnameend} (\bibinfo{year}{2000}):
  \emph{\bibinfo{title}{{Minimum description length induction, Bayesianism, and
  Kolmogorov complexity}}}.
\newblock {\sl \bibinfo{journal}{IEEE Transactions on Information Theory}}
  \bibinfo{volume}{46}(\bibinfo{number}{2}), pp. \bibinfo{pages}{446--464},
  \doi{10.1109/18.825807}.

\bibitemdeclare{article}{Walters1989}
\bibitem{Walters1989}
\bibinfo{author}{R.~F.C. \surnamestart Walters\surnameend}
  (\bibinfo{year}{1989}): \emph{\bibinfo{title}{{The free category with
  products on a multigraph}}}.
\newblock {\sl \bibinfo{journal}{Journal of Pure and Applied Algebra}}
  \bibinfo{volume}{62}(\bibinfo{number}{2}), pp. \bibinfo{pages}{205--210},
  \doi{10.1016/0022-4049(89)90152-7}.

\bibitemdeclare{inproceedings}{Webb2018}
\bibitem{Webb2018}
\bibinfo{author}{Stefan \surnamestart Webb\surnameend}, \bibinfo{author}{Adam
  \surnamestart Goli\'{n}ski\surnameend}, \bibinfo{author}{Robert \surnamestart
  Zinkov\surnameend}, \bibinfo{author}{N.~\surnamestart Siddharth\surnameend},
  \bibinfo{author}{Tom \surnamestart Rainforth\surnameend},
  \bibinfo{author}{Yee~Whye \surnamestart Teh\surnameend} \&
  \bibinfo{author}{Frank \surnamestart Wood\surnameend} (\bibinfo{year}{2018}):
  \emph{\bibinfo{title}{Faithful Inversion of Generative Models for Effective
  Amortized Inference}}.
\newblock In: {\sl \bibinfo{booktitle}{Proceedings of the 32nd International
  Conference on Neural Information Processing Systems}},
  \bibinfo{series}{NIPS'18}, \bibinfo{publisher}{Curran Associates Inc.},
  \bibinfo{address}{Red Hook, NY, USA}, p. \bibinfo{pages}{3074–3084}.

\bibitemdeclare{inproceedings}{Xu2019}
\bibitem{Xu2019}
\bibinfo{author}{Kai \surnamestart Xu\surnameend}, \bibinfo{author}{Akash
  \surnamestart Srivastava\surnameend} \& \bibinfo{author}{Charles
  \surnamestart Sutton\surnameend} (\bibinfo{year}{2019}):
  \emph{\bibinfo{title}{Variational Russian Roulette for Deep {B}ayesian
  Nonparametrics}}.
\newblock In \bibinfo{editor}{Kamalika \surnamestart Chaudhuri\surnameend} \&
  \bibinfo{editor}{Ruslan \surnamestart Salakhutdinov\surnameend}, editors:
  {\sl \bibinfo{booktitle}{Proceedings of the 36th International Conference on
  Machine Learning}}, {\sl \bibinfo{series}{Proceedings of Machine Learning
  Research}}~\bibinfo{volume}{97}, \bibinfo{publisher}{PMLR}, pp.
  \bibinfo{pages}{6963--6972}.
\newblock \urlprefix\url{https://proceedings.mlr.press/v97/xu19e.html}.

\bibitemdeclare{inproceedings}{Zhao2017}
\bibitem{Zhao2017}
\bibinfo{author}{Shengjia \surnamestart Zhao\surnameend},
  \bibinfo{author}{Jiaming \surnamestart Song\surnameend} \&
  \bibinfo{author}{Stefano \surnamestart Ermon\surnameend}
  (\bibinfo{year}{2017}): \emph{\bibinfo{title}{{Learning Hierarchical Features
  from Generative Models}}}.
\newblock In: {\sl \bibinfo{booktitle}{International Conference on Machine
  Learning}}.
\newblock \urlprefix\url{http://arxiv.org/abs/1702.08396}.

\bibitemdeclare{inproceedings}{Zhou2021}
\bibitem{Zhou2021}
\bibinfo{author}{Yanli \surnamestart Zhou\surnameend} \&
  \bibinfo{author}{Brenden~M. \surnamestart Lake\surnameend}
  (\bibinfo{year}{2021}): \emph{\bibinfo{title}{{Flexible Compositional
  Learning of Structured Visual Concepts}}}.
\newblock In: {\sl \bibinfo{booktitle}{Proceedings of the 43rd Annual
  Conference of the Cognitive Science Society}}.
\newblock \urlprefix\url{http://arxiv.org/abs/2105.09848}.

\end{thebibliography}

\appendix
\newpage

\section{Estimation of log-likelihood by importance weighting}
\label{app:importanceweighting}
In approximate inference techniques based on importance weighting, we sample latent variables from a proposal density $q_\phi$ and then score them with an importance weight
\begin{align}
    \label{eq:weight}
    w_{\theta,\phi} &= \frac{p_\theta(x, z, f, \beta, W; \Phi, G)}{q_\phi(z, f, \beta, W \mid x)}
\end{align}
equal to the ratio of the proposal joint density over the latent variables and the generative joint density over all variables. This weighting adjusts for the bias of the proposal density, relative to the normalized generative joint density (that is, the Bayesian inverse or posterior distribution).

\begin{lemma}[Importance weighting provides an estimator of the marginal density]
\label{lem:importance}
Given the generative and proposal joint distributions above, the expectation of the importance weights equals the marginal density of the observation
\begin{align}
    \label{eq:importance}
    \expect{q_\phi(z, f, \beta, W \mid x; \Phi)}{w_{\theta,\phi}} &= p_\theta(x; \Phi, G).
\end{align}
Finite samples approximating this expectation therefore provide Monte Carlo estimators of the analytically intractable marginal density:
\begin{align*}
    p_\theta(x; \Phi, G) &\approx \frac{1}{K} \sum_{k=1}^{K} w^{k}_{\theta,\phi} \text{ for } w^{k}_{\theta,\phi} \sim q_\phi(z, f, \beta, W \mid x; \Phi).
\end{align*}
\end{lemma}
\begin{proof}
We begin with the definition of the expected importance weight
\begin{align*}
    \expect{q_\phi(z, f, \beta, W \mid x; \Phi)}{w_{\theta,\phi}} &= \expectint{q_\phi(z, f, \beta, W \mid x; \Phi)}{z, f, \beta, W}{w_{\theta,\phi}},
\end{align*}
and expand it to include the density ratio
\begin{align*}
    &= \expectint{q_\phi(z, f, \beta, W \mid x; \Phi)}{z, f, \beta, W}{\left[
        \frac{p_\theta(x, z, f, \beta, W; \Phi, G)}{q_\phi(z, f, \beta, W \mid x)}
    \right]}, \\
    &= \expectint{}{z, f, \beta, W}{
        \frac{
            \cancel{q_\phi(z, f, \beta, W \mid x; \Phi)}
            p_\theta(x, z, f, \beta, W; \Phi, G)
        }{\cancel{q_\phi(z, f, \beta, W \mid x)}}
    }, \\
    &= \expectint{p_\theta(x, z, f, \beta, W; \Phi, G)}{z, f, \beta, W}{}, \\
    \expect{q_\phi(z, f, \beta, W \mid x; \Phi)}{w_{\theta,\phi}} &= p_\theta(x; \Phi, G).
\end{align*}
\end{proof}
This quantity can be estimated by Monte Carlo techniques, obtaining the estimator
\begin{align*}
    Z_{\theta,\phi} &\approx \frac{1}{K} \sum_{k=1}^{K} w_{\theta, \phi} & \Hat{Z}_{\theta,\phi} &= \frac{1}{K} \sum_{k=1}^{K} w_{\theta, \phi}.
\end{align*}
Resampling proportionally to these weights can then provide us with an unbiased sample from the true posterior distribution \cite{Chopin2020}. Storing the actual weights in log-space typically provides better numerical stability, when coupled to a specialized \texttt{mean-exp} implementation. In our case, PyTorch provides such an implementation.

Taking the logarithm of both sides of Equation~\ref{eq:importance} yields an expression known as the log-evidence for the model
\begin{align*}
    \log \expect{q_\phi(z, f, \beta, W \mid x; \Phi)}{w_{\theta,\phi}} &= \log p_\theta(x; \Phi, G),
\end{align*}
and applying Jensen's Inequality yields a variational lower bound to the log-evidence
\begin{align*}
    \expect{q_\phi(z, f, \beta, W \mid x; \Phi)}{\log w_{\theta,\phi}} &\leq \log p_\theta(x; \Phi, G).
\end{align*}
We call the left-hand side of this inequality the Evidence Lower Bound (ELBO)
\begin{align}
    \label{eq:elbo}
    \mathcal{L}(\theta, \phi) &= \expect{q_\phi(z, f, \beta, W \mid x; \Phi)}{\log w_{\theta,\phi}} \leq \log p_\theta(x; \Phi, G).
\end{align}
Appendix~\ref{app:elbo} will provide an alternate derivation for the ELBO, showing that maximizing the ELBO minimizes the (exclusive) Kullback Lieblier divergence from the proposal to the posterior distribution.

\section{Derivation of the ELBO objective}
\label{app:elbo}

Any proposal like Equation~\ref{eq:jointproposal} can help sample from the posterior distribution by importance weighting, but we can obtain progressively better approximations by minimizing the Kullback Liebler divergence (see \cite{Murphy2012} for details) from the proposal to the true posterior
\begin{align*}
    \kl{q_\phi(z, f, \beta, W \mid x)}{p_\theta(z, f, \beta, W \mid x; \Phi, G)} &= \expect{
        q_\phi(z, f, \beta, W \mid x)
    }{
        \log \frac{
            q_\phi(z, f, \beta, W \mid x)
        }{
            p_\theta(z, f, \beta, W \mid x; \Phi, G)
        }
    }.
\end{align*}
Equation~\ref{eq:bayesinverse} shows the true posterior to be
\begin{align*}
    p_\theta(z, f, \beta, W \mid x; \Phi, G) &= \frac{
        p_\theta(x, z, f, \beta, W; \Phi, G)
    }{p_\theta(x; \Phi, G)},
\end{align*}
with the numerator being the joint distribution defined in Equation~\ref{eq:genjoint} above. Substituting the right-hand side of this equation into the definition of the divergence above
\begin{align*}
    \kl{q_\phi}{p_\theta} &= \expect{
        q_\phi(z, f, \beta, W \mid x)
    }{
        \log \frac{
            q_\phi(z, f, \beta, W \mid x) p_\theta(x; \Phi, G)
        }{
            p_\theta(x, z, f, \beta, W; \Phi, G)
        }
    }, \\
    &= \expect{
        q_\phi(z, f, \beta, W \mid x)
    }{
        \log q_\phi(z, f, \beta, W \mid x) + \log p_\theta(x; \Phi, G)
        - \log p_\theta(x, z, f, \beta, W; \Phi, G)
    }, \\
    &= \expect{
        q_\phi(z, f, \beta, W \mid x)
    }{
        \log q_\phi(z, f, \beta, W \mid x)
        - \log p_\theta(x, z, f, \beta, W; \Phi, G)
    } + \log p_\theta(x; \Phi, G), \\
    &= \expect{
        q_\phi(z, f, \beta, W \mid x)
    }{
        \log \frac{q_\phi(z, f, \beta, W \mid x)}{p_\theta(x, z, f, \beta, W; \Phi, G)}
    } + \log p_\theta(x; \Phi, G), \\
    &= \log p_\theta(x; \Phi, G) - \expect{
        q_\phi(z, f, \beta, W \mid x)
    }{
        \log \frac{p_\theta(x, z, f, \beta, W; \Phi, G)}{q_\phi(z, f, \beta, W \mid x)}
    }
\end{align*}
shows that the divergence decomposes into two components: the log model evidence and the negative expected log importance weight. The sum of the expected log weight and the divergence is the log model evidence
\begin{align*}
    \kl{q_\phi}{p_\theta} + \expect{
        q_\phi(z, f, \beta, W \mid x)
    }{
        \log \frac{p_\theta(x, z, f, \beta, W; \Phi, G)}{q_\phi(z, f, \beta, W \mid x)}
    } &= \log p_\theta(x; \Phi, G),
\end{align*}
so that the expected log weight itself
\begin{align*}
    \expect{
        q_\phi(z, f, \beta, W \mid x)
    }{
        \log \frac{p_\theta(x, z, f, \beta, W; \Phi, G)}{q_\phi(z, f, \beta, W \mid x)}
    } &= \log p_\theta(x; \Phi, G) - \kl{q_\phi}{p_\theta},
\end{align*}
equals the log evidence minus the divergence. Since the divergence itself is always non-negative
\begin{align*}
    \log p_\theta(x; \Phi, G) -\kl{q_\phi}{p_\theta} &\leq \log p_\theta(x; \Phi, G) \\
    \expect{
        q_\phi(z, f, \beta, W \mid x)
    }{
        \log \frac{p_\theta(x, z, f, \beta, W; \Phi, G)}{q_\phi(z, f, \beta, W \mid x)}
    } &\leq \log p_\theta(x; \Phi, G),
\end{align*}
the expected log-weight therefore provides a lower bound to the log-evidence. For this reason, we typically call it the Evidence Lower Bound (ELBO)
\begin{align*}
    \mathcal{L}(\theta, \phi) &= \mathbb{E}_{q_{\phi}}\left[ \log \frac{p_\theta(x, z, f, \beta, W; \Phi, G)}{q_\phi(z, f, \beta, W \mid x)} \right],
\end{align*}
and take it as an objective function to perform variational Bayesian inference \cite{Blei2016}. The derivation above implies that maximizing $\mathcal{L}$ will minimize the divergence between the tractable proposal density and the intractable true posterior, with $\mathcal{L}$ equalling the true model-evidence only if the divergence reaches zero.

\section{Training and evaluation details}
\label{app:training}
We parameterized the free category $\freeCat{Q}$ by generating input and output pairs from dimensionalities ranging from 4 to 196, picking powers of 2, and constructing the following build blocks for deep generative models:
\begin{enumerate}
    \item Fully-connected VAE decoder networks. Decoders mapping down to $\mathbb{R}^{196}$ were considered to decode image glimpses, and therefore used transposed convolutional layers to produce their outputs;
    \item Variational ``Ladder'' decoder and prior networks from Zhao~\cite{Zhao2017}'s work on learning hierarchical features, with a noise dimension of 2; and
    \item Spatial attention mechanisms from Rezende~\cite{Rezende2016} which map $\mathbb{R}^{14\times 14} \rightarrow \mathbb{R}^{28\times 28}$, sampling a spatial transformation code $z\in \mathbb{R}^3$ from a learned Gaussian prior.
\end{enumerate}
Attaching the likelihood to the generator morphism required sampling the free category prior via the two-part wiring diagram
\begin{align*}
    f_{\theta} &: \mathbf{QBS}(I, \prob{(\mathbb{R}^{28\times 28} \times \tau)}) &
    \ell_{\sigma} &: \mathbf{QBS}(\mathbb{R}^{28\times 28}, \prob{(\mathbb{R}^{28\times 28} \times (\mathtt{x}: \mathbb{R}^{28\times 28}))}) \\
    f_{\theta} \fatsemi_{\prob{}} \ell_{\sigma} &: \mathbf{QBS}(I, \prob{(\mathbb{R}^{28\times 28} \times (\mathtt{x}: \mathbb{R}^{28\times 28}))}).
\end{align*}
$\mathcal{L}$ and its gradients were approximated by Monte Carlo sampling; Pyro~\cite{bingham2019pyro}'s \texttt{TraceGraphELBO} class provided gradient estimators~\cite{Mnih2014,Schulman2015}. We trained this model for 1200 epochs, with a learning rate $\eta=10^{-3}$. At test time we substitute a Bernoulli likelihood for the Gaussian, to ``compare like to like'' with other models in the literature.

\section{Dataset history and selection of competing models}
\label{app:competing}

Lake~\cite{Lake2015,Lake2019b} and colleagues proposed the Omniglot dataset to challenge the machine learning community to achieve human-like concept learning by learning a single generative model from very few examples; the Omniglot challenge requires that a model be usable for classification, latent feature recognition, concept generation from a type, and exemplar generation of a concept. The deep generative models research community has focused on producing models capable of few-shot reconstruction of unseen characters. Rezende~\cite{Rezende2016} and Hewitt~\cite{Hewitt2018} fixed as constant the model architecture, attempting to account for the compositional structure in the data with static dimensionality. In contrast, He~\cite{He2019} and Feinman~\cite{Feinman2020} jointly learned the model structure alongside inferring the posterior distribution over the latent variables and reconstructing the data.

\end{document}